\documentclass[lettersize,journal]{IEEEtran}
\usepackage{amsmath,amsfonts,mathtools}
\usepackage{array}
\usepackage{textcomp}
\usepackage{stfloats}
\usepackage{url}
\usepackage{verbatim}
\usepackage{graphicx}
\usepackage{cite}
\usepackage[T1]{fontenc}

\usepackage{xcolor,soul,framed} 

\colorlet{shadecolor}{yellow}

\graphicspath{{../pdf/}{../jpeg/}}
\DeclareGraphicsExtensions{.pdf,.jpeg,.png}

\usepackage{mdwmath}
\usepackage[square,numbers,sort&compress]{natbib}
\usepackage{chngcntr}
\usepackage{multirow}

\usepackage{wasysym}
\usepackage{cite}
\usepackage{caption}
\usepackage{subcaption}

\usepackage{amssymb}
\usepackage{booktabs}
\usepackage{makecell}
\usepackage{setspace}
\usepackage{enumerate}
\usepackage{pifont}
\usepackage{ragged2e}
\usepackage{hhline}
\usepackage{colortbl}
\usepackage{mdframed}
\usepackage{balance}
\usepackage{marvosym}
\usepackage{booktabs}
\usepackage{hyperref}
\usepackage[capitalise]{cleveref}

\usepackage[linesnumbered,ruled, vlined]{algorithm2e}
\usepackage{algpseudocode}  

\crefname{algocf}{Algorithm}{Algorithms}
\Crefname{algocf}{Algorithm}{Algorithms}

\begin{document}

\title{TrustGeoGen: Formal-Verified Data Engine for Trustworthy Multi-modal Geometric Problem Solving}

\author{Daocheng Fu, Jianlong Chen, Renqiu Xia, Zijun Chen, Qi Liu, Yuan Feng, Hongbin Zhou, Renrui Zhang, Shiyang Feng, Peng Gao, Hongyuan Zha, Junchi Yan, Botian Shi, Yu Qiao, Bo Zhang
\thanks{Daocheng Fu, Jianlong Chen, Renqiu Xia and Zijun Chen contribute equally to this work.}
\thanks{Corresponding authors: Bo Zhang (E-mail: zhangbo@pjlab.org.cn) and Renqiu Xia (E-mail: xiarenqiu@sjtu.edu.cn).}
\thanks{Daocheng Fu is with College of Computer Science and Artificial Intelligence, Fudan University.}
\thanks{Daocheng Fu, Hongbin Zhou, Renrui Zhang, Shiyang Feng, Peng Gao, Botian Shi, Yu Qiao and Bo Zhang are with Shanghai Artificial Intelligence Laboratory, Shanghai 200232, China.}
\thanks{Jianlong Chen and Hongyuan Zha are with The Chinese University of Hong Kong, Shenzhen.}
\thanks{Renqiu Xia, Zijun Chen, Qi Liu, Yuan Feng and Junchi Yan are School of Artificial Intelligence, Shanghai Jiao Tong University.}
} 

\maketitle

\newcommand{\method}{TrustGeoGen}
\renewcommand{\algorithmicrequire}{\textbf{Input:}}  
\renewcommand{\algorithmicensure}{\textbf{Output:}} 

\begin{abstract}


\justifying{Geometric problem solving (GPS) requires precise multimodal understanding and rigorous, step-by-step logical reasoning. However, developing capable Multimodal Large Language Models (MLLMs) for GPS is heavily bottlenecked by the scarcity of high-quality, verifiable data. Existing data acquisition paradigms either suffer from modality incompleteness and unverified logical gaps ("leaps-of-faith"), or rely on formal engines that generate rigid, structurally homogeneous data, failing to produce high-difficulty problems or foster genuine natural-language reasoning. To overcome these limitations, we introduce TrustGeoGen, an autonomous and formalized geometric data generation engine. TrustGeoGen strictly guarantees reasoning trustworthiness through formal verification while generating multimodally integrated data, including premises, visual diagrams, and solutions. To systematically scale problem difficulty, we incorporates difficulty-aware filtering and iterative bootstrapping mechanism. Furthermore, we propose "connection thinking" to bridge the semantic gap between rigid formal logic and fluent human-like reasoning, ensuring coherent logical transitions. We also introduce the GeoExplore family of sampling algorithms to extract diverse problem-solving trajectories based on various thinking templates. Extensive experiments demonstrate that training models on our synthesized dataset, GeoTrust, substantially enhances deep geometric reasoning capabilities and yields significant performance gains across out-of-distribution (OOD) benchmarks, including GeoQA, Geometry3K, and OlympiadBench.
}

\end{abstract}

\begin{IEEEkeywords}
Data Engine, Geometry, Multi-modal
\end{IEEEkeywords}

\section{Introduction}
\label{sec:intro}

Geometric problem solving (GPS)~\cite{geox/Xia2024GeoXGP,peng-etal-2023-geodrl,wu2024gps} is a representative task for mathematical reasoning. It requires models to jointly interpret diagrams, understand textual conditions, and produce logically valid multi-step solutions, where each conclusion should be justified by prior statements~\cite{feng2025geobench}. As a result, GPS is a challenging benchmark for trustworthy multimodal reasoning: a correct final answer alone is insufficient if the intermediate reasoning is incomplete or unjustified. Progress on this task therefore depends on multimodal data that aligns diagrams, language, and verifiable reasoning steps.

Recent multimodal large language models (MLLMs)~\citep{math-llava/shi2024math,mathvista/lumathvista,peng2024chimera,qwen-vl/bai2023qwen,openai2024gpt4o,g-llava/gao2023g,chen2024far,lu2024deepseek,geox/Xia2024GeoXGP,wu2024deepseekvl2mixtureofexpertsvisionlanguagemodels,internlmxcomposer2,internlmxcomposer2_4khd} have shown encouraging progress on elementary geometry, while formalized expert systems~\citep{unigeo_geoformer/chen2022unigeo,unimath/liang2023unimath,pgpsnet/zhang2023multi,inter-gps/lu2021inter} and specialized solvers~\citep{alphageometry/Trinh2024,alphageometry2/chervonyi2025,tonggeometry/Zhang2024ProposingAS} have demonstrated strong potential on more challenging problems. However, existing pipelines still do not provide models with sufficiently coherent multimodal reasoning supervision. In particular, when language models are used mainly to interact with formal solvers, the resulting data may be formally correct but linguistically rigid, making it difficult for models to learn natural, well-connected reasoning explanations. This limitation highlights the need for training data that is both formally verifiable and natural-language coherent.

As shown in Table~\ref{tab:datasets_comparison}, current multimodal geometry data mainly comes from two paradigms: annotating existing problems~\cite{geoqa_ngs/chen2021geoqa,dpe-gps/cao2022augmented,zhang2024geoeval,leonardi2024geosence} and synthesizing data with formal engines~\cite{inter-gps/lu2021inter,pgpsnet/zhang2023multi}. Annotation-based approaches rely on textbook problems or curated datasets, with solutions written manually or assisted by LLMs. While practical, they are constrained by the source data and often suffer from incomplete modalities or missing intermediate steps, especially for difficult problems. Formal synthesis, in contrast, can produce stepwise verifiable reasoning at much lower annotation cost. Yet existing synthetic pipelines typically lack systematic difficulty control, and the generated explanations are often template-like, weakly connected across steps, and limited in reasoning diversity. Consequently, existing datasets still leave substantial room for improvement in completeness, verifiability, and reasoning quality.

\begin{table*}[htbp]
\centering	
    \caption
	{
        Comparison of various geometric problem-solving datasets and benchmarks.
        Modality abbreviations: I for Image, FS for Formal Solution, NS for Natural Solution, A for Answer.
        Thinking Template abbreviations: D for Deductive, BT for Backtrack, MS for Multi-solution.
	}
    \vspace{-5pt}
	\resizebox{\textwidth}{!}{
	\begin{tabular}{l | l | l | l | l | l} 
	\toprule	 	
	\textbf{Datasets \& Benchmarks} & \textbf{Size} & \textbf{Modality} & \textbf{Level} & \textbf{Thinking Template} & \textbf{Labeling Method} \\
     \midrule
    GeoQA~\cite{geoqa_ngs/chen2021geoqa} & 4998 & I, NS, A & Middle School & D & Human Annotation \\
    GeoQA+~\cite{dpe-gps/cao2022augmented} & 7528 & I, NS, A & Middle School & D & Human Annotation \\
    Geometry3K~\cite{inter-gps/lu2021inter} & 3002 & I, FS, A & Middle School & D & Step-wise Formal Verified \\
    PGPS9K~\cite{pgpsnet/zhang2023multi} & 9022 & I, FS, A & Middle School & D & Step-wise Formal Verified \\
    UniGeo~\cite{unigeo_geoformer/chen2022unigeo} & 9543 & I, NS, A & Middle School & D & From Source File \\
    GeoEval~\cite{zhang2024geoeval} & 5050 & I, A & Middle \& High School & - & LLM Labeling \\
    MathVista~\cite{mathvista/lumathvista} & 1320 * & I, A & Middle School & - & From Source File \\
    MathVerse~\cite{zhang2024mathverse} & 1745 * & I, NS, A & High School & - & LLM Labeling \\
    GeoSense~\cite{leonardi2024geosence} & 1789 & I, NS, A & Synthetic & D & LLM Labeling \& Human Annotation \\
    OlympiadBench~\cite{he2024olympiadbench} & 8476 & I, NS, A & Olympiad-level & D & From Source File \\
    MAVIS~\cite{mavis/zhang2024mavis} & 588K & I, NS, A & Synthetic & D & LLM Labeling \\
    G-LLAVA~\cite{g-llava/gao2023g} & 170K & I, NS, A & Synthetic & D & LLM Lableling \\
    AutoGeo~\cite{huang2025autogeo} & 100K & I, FS, NS, A & Synthetic & D & Step-wise Formal Verified \\
    \midrule
    TrustGeoGen & 200K & I, FS, NS, A & Synthetic & D, MS, BT & Step-wise Formal Verified \\
    GeoTrust-train & 2342 & I, FS, NS, A & Synthetic & D, MS, BT & Step-wise Formal Verified \\
    GeoTrust-test & 240 & I, A & Synthetic & - & Step-wise Formal Verified \\
    \bottomrule
	\end{tabular}
	}
	\label{tab:datasets_comparison}
    \par 
    \raggedright 
    \footnotesize 
     \hspace{0.5em}* These values are not the total count but the count of the GPS category within the dataset.
\end{table*}

To address these issues, we propose \textbf{TrustGeoGen}, a formally verified generation engine for multimodal geometric data. TrustGeoGen consists of four components: (1) a \textit{Constructor} that generates geometric premises and corresponding diagrams under explicit constraints; (2) a \textit{Reasoner} that expands formally valid reasoning graphs through rule-based verification; (3) a \textit{Sampler} that uses \textit{GeoExplore} to extract high-quality reasoning paths and formulate problem-solution pairs; and (4) a \textit{Translator} that converts formal derivations into fluent natural-language explanations. To generate more challenging and informative data, TrustGeoGen further incorporates difficulty-aware filtering and iterative bootstrapping. In addition, we introduce \emph{connection thinking}, which bridges the semantic gap between rigid formal logic and fluent human-like reasoning, ensuring coherent logical transitions, and use the \textit{GeoExplore} family of sampling algorithms to extract diverse problem-solving trajectories based on various thinking templates.

Experiments show that TrustGeoGen can generate challenging geometry data and that training on the resulting dataset consistently improves model performance. The proposed translation strategy helps narrow the gap between formally verified derivations and human-readable reasoning, while the \textit{GeoExplore} sampling algorithms further improve reasoning diversity. Models trained on the synthesized dataset, GeoTrust, achieve substantial gains on out-of-distribution benchmarks, including GeoQA~\citep{geoqa_ngs/chen2021geoqa}, Geometry3K~\cite{inter-gps/lu2021inter}, and OlympiadBench~\cite{he2024olympiadbench}, demonstrating strong generalization to unseen geometry problems. Our main contributions are as follows:

\begin{itemize}
    \item We propose \textbf{TrustGeoGen}, a generation engine for geometric data that produces integrated multimodal outputs, including diagrams, formal descriptions, natural-language explanations, questions, and solutions, with formally verifiable intermediate reasoning steps.
    \item We introduce \emph{connection thinking} to improve the coherence between formal derivations and natural-language explanations, and propose the \textit{GeoExplore} sampling family to increase the diversity of generated reasoning paths.
    \item Extensive experiments show that data generated by \textbf{TrustGeoGen} is challenging, verifiable, and effective for improving geometric problem-solving performance, with strong transfer to out-of-distribution benchmarks.
\end{itemize}
\section{Related Work}

\subsection{Advances in Multimodal Large Language Model} 

Recent advances in large language models (LLMs) have demonstrated remarkable progress in linguistic intelligence, achieving human-level capabilities in various applications~\citep{chatgpt/ouyang2022training,llama/touvron2023llama,llama2/touvron2023llama,internlm/team2023internlm}. Building upon this foundation, the research community has focused on extending these text-based architectures to process visual information, leading to the emergence of sophisticated multimodal large language models (MLLMs)~\citep{qwen-vl/bai2023qwen, gpt-4v/achiam2023gpt, reid2024gemini}.
These integrated systems typically employ modality alignment mechanisms, such as Q-former~\citep{blip-2/li2023blip} and linear layers~\citep{llava/liu2024visual}, to establish connections between visual representations and text embeddings. Recently, an important application of MLLMs is their ability to be regarded as multimodal agents that interact with scientific environments~\citep{team2025novelseek,zhang2025origene,gottweis2025towards}, thereby generating valuable scientific discoveries and driving productivity advancements. Although current MLLMs show strong performance in standard vision-language benchmarks~\citep{fu2023mme,xia2024docgenome,structchart/xia2023structchart,mathvista/lumathvista,jiang2025mme}, their effectiveness decreases significantly when processing mathematical visualizations that require reasoning. Recent specialized approaches address this limitation through targeted training strategies. For example, geometric reasoning capabilities have been enhanced through domain-specific datasets containing annotated diagrams~\citep{g-llava/gao2023g,mavis/zhang2024mavis}.

\subsection{Geometric Problem Solving} 
As a challenging task, geometric problem solving requires understanding diagrams, interpreting symbols, and performing complex reasoning. While MLLMs have shown remarkable proficiency in human-level reasoning and generation capabilities, they struggle with automatic geometric problem solving due to their pre-training on natural images and texts, as well as the lack of automated verification in the problem-solving process. To address this limitation, researchers have proposed various approaches to improve the understanding of geometric images and corpora, including unimodal pre-training, vision-language alignment, and visual instruction tuning~\citep{geox/Xia2024GeoXGP,mavis/zhang2024mavis,g-llava/gao2023g,geoqa_ngs/chen2021geoqa,unigeo_geoformer/chen2022unigeo,pgpsnet/zhang2023multi,pgdp5k/hao2022pgdp5k,jiang2025mme}. Another major approach involves the use of formalized solvers to tackle geometric problems. While these solvers~\citep{alphageometry/Trinh2024,alphageometry2/chervonyi2025,sicca2024newclid} are capable of addressing challenges at the level of the international mathematical olympiad (IMO), they require precise modeling of geometric problems to fit the solver's language. This necessity introduces significant obstacles in terms of practical application and generalization, limiting the usability in real-world scenarios.

\subsection{Datasets and Benchmarks}

High-quality training data is crucial for enhancing the geometry problem-solving capabilities of MLLMs. As summarized in~\cref{tab:datasets_comparison}, there are three common data construction approaches. The first approach involves filtering real-world data and then annotating it manually~\cite{geoqa_ngs/chen2021geoqa, dpe-gps/cao2022augmented, unigeo_geoformer/chen2022unigeo, mathvista/lumathvista, he2024olympiadbench}. While this method yields high-quality multi-modal reasoning data, its scalability is limited by the finite data pool and the intensive labor required. Furthermore, as problem complexity increases, the difficulty of annotation rises sharply, typically constraining the resulting problems to a lower difficulty range. The second approach utilizes formal engines for data generation~\cite{inter-gps/lu2021inter, pgpsnet/zhang2023multi}. This allows for the rapid creation of large-scale datasets with verifiably correct reasoning processes. However, the quality of the data is inherently tied to the sophistication of the formal engine. A significant gap often exists between the synthesized logic and authentic human reasoning, leaving considerable room for improvement in enhancing the model's natural language inference abilities. Finally, the third method employs LLMs to synthesize reasoning data~\cite{mavis/zhang2024mavis, g-llava/gao2023g}. This method is highly efficient, and the generated reasoning steps more closely resemble natural language. Its primary drawback, however, is that the outputs are unverifiable; they may contain erroneous intermediate steps that could mislead or corrupt the model's reasoning capabilities.

\section{Methodology}
The data engine TrustGeoGen serves as the core for constructing high-quality geometric reasoning datasets, enabling the generation of complex geometric scenes and reasoning paths to support robust and scalable inference.

\begin{figure*}[ht]
\vspace{-8pt}
\centering
\includegraphics[width=0.99\linewidth]{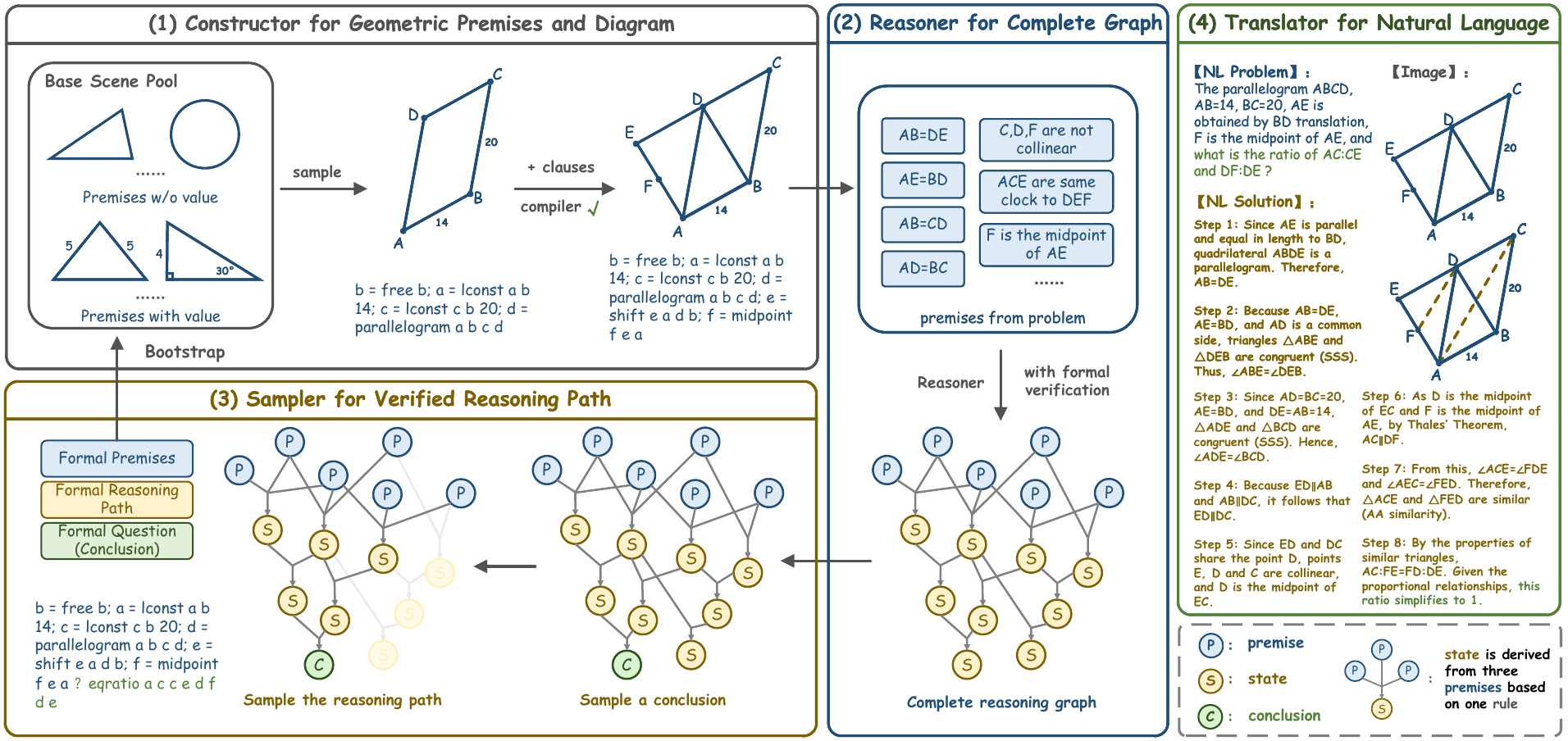}
\caption{Overview of TrustGeoGen: (1) \textbf{Constructor} builds geometric premises and visual diagrams; (2) \textbf{Reasoner} generates a formally validated reasoning graph; (3) \textbf{Sampler} utilizes \textit{GeoExplore} algorithms to select conclusions and trace diverse reasoning paths; (4) \textbf{Translator} converts formal logic into fluent natural language using ``connection thinking''. A bootstrap mechanism amplifies problem difficulty iteratively.}
\label{fig:framework}
\end{figure*}

\subsection{Data Engine}
In the process of generating multi-modal geometric reasoning data, it employs a rule-based data construction method augmented by a geometric verifier to synthesize high-quality datasets. This approach ensures that the generated data are accurate and controllable. As shown in ~\cref{fig:framework}, our framework consists of four components: Constructor, Reasoner, Sampler, and Translator.

\textbf{Constructor} is responsible for generating a geometric scene for a given problem. The foundational components of this process are constructions ($C$) and statements ($s$). A geometric scene is described by a sequence of constructions, where each construction $C_i$ generates one or more statements, such as $\{s_i^1, s_i^2, ...\}$, that define the relationships between geometric elements. Before a new construction can be added to the scene, the existing statements must satisfy its preconditions. For instance, to apply the construction "draw a perpendicular from point $A$ to line $BC$, with $D$ as the foot," the precondition "points $A$, $B$, and $C$ are not collinear" must hold true. Conversely, the successful application of a construction yields new statements. For example, the construction "create an isosceles triangle $ABC$" introduces the statements "$AB = AC$" and "$\angle ABC = \angle ACB$". Following this paradigm, a complete geometric scene, described by $\{C_1, C_2, ..., C_n\}$, can be built incrementally from scratch. This process concurrently generates the scene's definitive set of initial statements, $S_0 = \{s_1^1, s_1^2, ..., s_n^1, s_n^2, ...\}$, which is the total collection of all generated statements.

To ensure the generated scenes are both geometrically significant and numerically instantiated, we have hand-crafted a library of 46 base scene generation functions. These functions produce primitive geometric configurations populated with random numerical values. As shown in~\cref{fig:framework}, the \textbf{Constructor} initiates the process by randomly selecting one of these functions to create an initial, numerically-defined scene. Subsequently, it iteratively adds further constructions, contingent on their preconditions being met, until a target scene is fully formed. This scene is then passed to a geometric compiler for validation, which checks for the consistency of all geometric relations and constraints. Ultimately, once a valid geometric scene is successfully constructed, its complete set of initial statements, $S_0$, can be obtained.

\begin{figure}[tb!]
\centering
\includegraphics[width=0.99\linewidth]{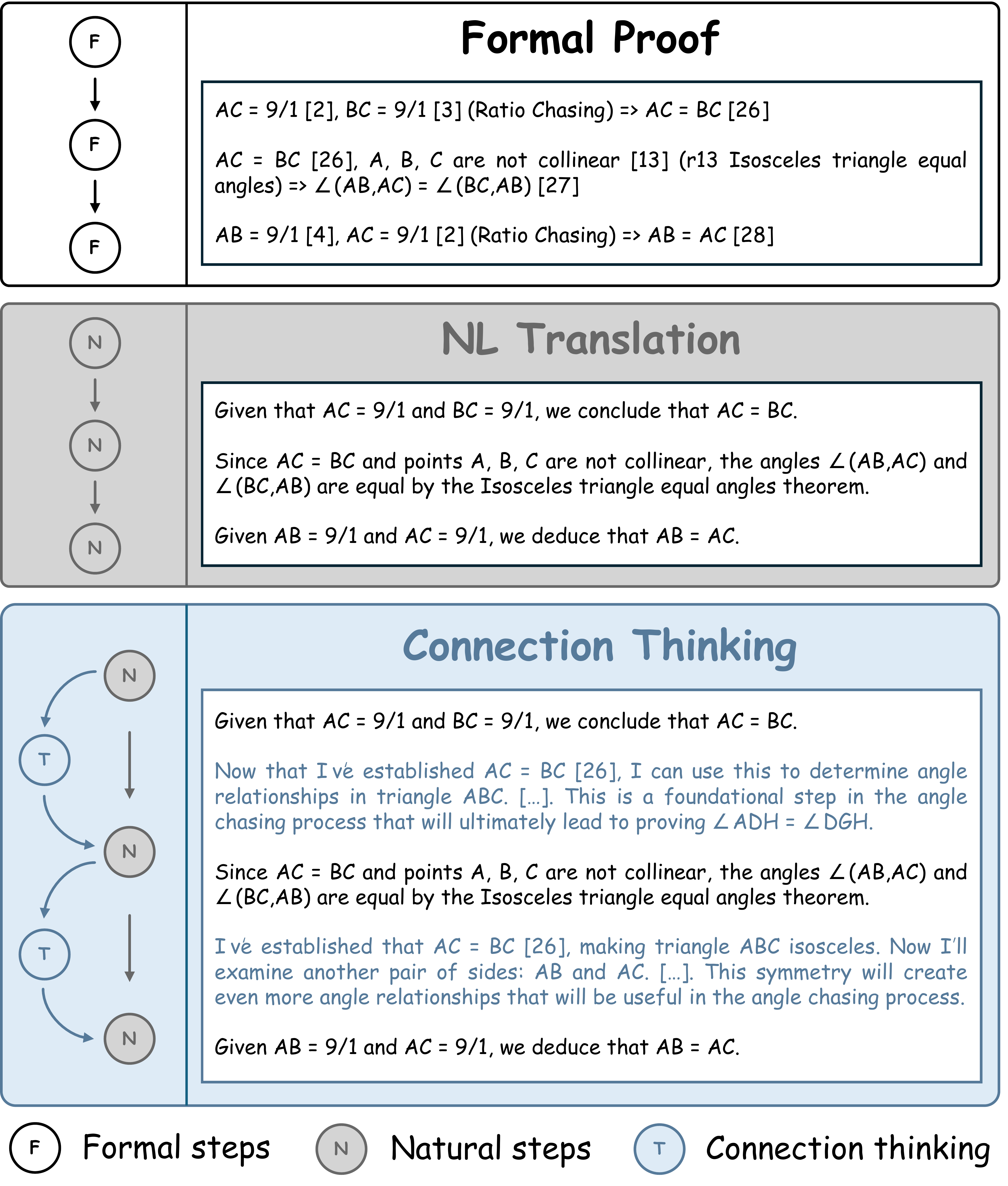}
\caption{Simple illustration of how Translator works}
\label{fig:translator_illus}
\vspace{-5pt}
\end{figure}

\textbf{Reasoner} leverages a predefined set of geometric theorems to infer new statements from premises generated by the \textit{constructor}, formulating a reasoning graph:
\begin{equation}
    \mathcal{G} = (S, S_0, R, \hookrightarrow),
\end{equation}
\begin{itemize}
    \item {\small $S$} is the set of statements, where each statement {\small $s \in S$} represents a derived conclusion or fact in the reasoning process.
    \item {\small $S_0 \subseteq S$} is the set of initial statements generated by the \textit{constructor}.
    \item {\small $R$} is the set of rules, where each rule $r \in R$ defines a logical inference relationship between statements.
    \item {\small $\hookrightarrow \  \subseteq S \times R \times S$ = $\{ (S_r,r,s') \mid S_r \subset S, r \in R, s' \in S \}$} is the statement transition relation. A transition {\small $S_r \xhookrightarrow{r} s'$} indicates statement {\small $s'$} is derived from {\small $s$ by applying rule $r$}.
\end{itemize}
\vspace{0.3em}

Reasoning begins with the initial statement {\small$S_0$} (given premises from the \textbf{constructor}), forming the root nodes of the graph {\small$\mathcal{G}$}. 
The graph expands by applying inference rules to existing statements. For every statement set {\small$S_r \subset S$ and rule $r \in R$, if $r$} can be applied to {\small$S_r$} to derive a new statement {\small$s'$} (formally {\small$S_r \xhookrightarrow{r} s'$}), the reasoner performs two atomic updates: 1) Add the new statement {\small$S_r$} to {\small$S$} and 2) Add the edges {\small$(S_r, s')$} to the transition relation {\small$\hookrightarrow$ with rule $r$}. Finally,  the complete graph is the smallest closure satisfying:

\begin{equation}
    S = S_0 \cup \left\{ s' \mid \forall S_r \subset S, \, r \in R, \, S_r \xhookrightarrow{r} s' \right\}.
\end{equation}

\begin{algorithm}[t]
  \SetKwInput{KwRequire}{Require}
  \SetKwInput{KwEnsure}{Ensure}

  \caption{GeoExplore}
  \label{algorithm:sampler}
  
  \KwRequire{Reasoning graph $\mathcal{G} = (S, S_0, R, \hookrightarrow)$, target statement $s_n$, threshold $\tau_l$, threshold $\tau_r$}
  \KwEnsure{Filtered reasoning path $\mathcal{P}$}
  
  \textbf{Initialize} $\mathcal{P} \gets \emptyset$\;
  $s \gets s_n$\;
  
  \While{$s \notin S_0$}{
    Identify rule $r_s \in R$ and parent statement set $S_{i-1} \subset S$ such that $S_{i-1} \xhookrightarrow{r_s} s$\;
    Append transition $(S_{i-1}, r_s, s)$ to $\mathcal{P}$\;
    $s \gets S_{i-1}$\;
  }
  
  Assess path length: $L \gets |\mathcal{P}|$\;
  Compute premises ratio: $R_P \gets |S_P| / |S_0|$, where $S_P \subseteq \mathcal{P}$\;
  
  \If{$L \geq \tau_l$ \textbf{and} $R_P \geq \tau_r$}{
    \KwRet{$\mathcal{P}$}\;  
  }
  \Else{
    \textbf{Discard} $\mathcal{P}$\;
  }
\end{algorithm}

\textbf{Sampler} operates on the reasoning graph {$\mathcal{G}$} by sampling a target statement $s_n$ and subsequently searching for its path:
\begin{equation}
\label{eq:sampler}
    \mathcal{P} = \{(S_{i-1}, r_s, s) \mid \forall s \in S_i, S_{i-1} \xhookrightarrow{r_s} s , i = n, \dots, 1\},
\end{equation}
where each triplet {\small$(S_i, r_s, s)$} denotes a specific transition during the reasoning process, derived by applying rule {\small$r_s$} to statement set {\small$S_{i-1}$} to produce statement {\small$s$}. Any statement {\small$s_n$} within the statement set {\small$S$} can be treated as a problem to solve in order to derive a reasoning path {\small$\mathcal{P}$}. The sampler is initialized with a statement {\small$s \in S_n$} to retrieve the corresponding transition rule {\small$r_s$} and the upstream dependent statements {\small$S_i$}. This process is repeated iteratively until the statement set {\small$S_i$} consists entirely of initial premises {\small$S_0$}. At that point, the full reasoning path has been successfully constructed. Notably, during the graph construction process in the \textbf{Reasoner}, once a statement {\small$s$} is derived through rule {\small$r_s$}, no additional statement transition relations are appended to {\small$s$}. Consequently, the \textit{Sampler} is always able to identify a unique inference rule {\small$r_s$} corresponding to any statement {\small$s$}, thereby guaranteeing the absence of cycles in the reasoning path.

To ensure the quality of the reasoning paths, the \textbf{Sampler} employs two metrics to filter out unsuitable paths. The first metric evaluates the length of the reasoning path {\small$\mathcal{P}$} against a predefined threshold {\small$\tau_l$}. If the length of {\small$\mathcal{P}$ exceeds $\tau_l$}, the path is considered difficult enough. Formally, this condition is expressed as {\small$|\mathcal{P}| \ge \tau_l$}, where {\small$|\mathcal{P}|$} denotes the number of transitions in the reasoning path {\small$\mathcal{P}$}. The second metric assesses the ratio of the statements in the reasoning path {\small$\mathcal{P}$} that are derived from the initial premises set {\small$S_0$}. If this ratio exceeds a predefined threshold {\small$\tau_r$}, the path is deemed sufficient enough. This condition is formally defined as {\small$\frac{|S_P|}{|S_0|} \ge \tau_r$}, where {\small$|S_P|$} represents the number of initial premises used in the reasoning path {\small$\mathcal{P}$}, and {\small$|S_0|$} is the number of all initial premises. By applying these two metrics, the \textbf{Sampler} ensures that only high-quality reasoning paths are retained, thereby maintaining the integrity and usefulness of the generated data. The detailed procedure of the aforementioned reasoning path exploration and filtering algorithm, \textit{GeoExplore}, is presented in ~\cref{algorithm:sampler}.

\begin{algorithm}[tb!]
  \SetKwInput{KwRequire}{Require}
  \SetKwInput{KwEnsure}{Ensure}

  \caption{GeoExplore-M}
  \label{algorithm:multi-sol}

  \KwRequire{Reasoning graph $\mathcal{G} = (S, S_0, R, \hookrightarrow)$, target statement $s_n$, threshold $\tau_l$, threshold $\tau_r$, number of searches $n$}
  \KwEnsure{Set of filtered reasoning paths $\mathcal{P}_{\text{set}}$}

  \textbf{Initialize} $\mathcal{P}_{\text{set}} \gets \emptyset$\;
  \textbf{Initialize} $\text{used\_options} \gets \emptyset$\; 
  
  \While{True}{
    \textbf{Initialize} $\mathcal{P} \gets \emptyset$\;
    $s \gets s_n$\;
    
    \While{$s \notin S_0$}{
      Identify set of rules $\mathcal{R}_s \subseteq R$ and corresponding parent statement sets $\mathcal{S}^m_{i-1} \subseteq S$ such that $\forall r_s \in \mathcal{R}_s, \exists S_{i-1} \in \mathcal{S}^m_{i-1}: S_{i-1} \xhookrightarrow{r_s} s$\;
      Select $r_s \in \mathcal{R}_s$ and $S_{i-1} \in \mathcal{S}^m_{i-1}$ (choose different options in different searches)\;
      $\text{used\_options} \gets \text{used\_options} \cup \{(r_s, S_{i-1})\}$\;
      
      \If{$(S_{i-1}, r_s, s) \notin \mathcal{P}$}{
        $\mathcal{P} \gets \mathcal{P} \cup (S_{i-1}, r_s, s)$\;
      }
      $s \gets S_{i-1}$\;
    }
    
    Assess path length: $L \gets |\mathcal{P}|$\;
    Compute premises ratio: $R_P \gets |S_P| / |S_0|$, where $S_P \subseteq \mathcal{P}$\;
    
    \If{$L \geq \tau_l$ \textbf{and} $R_P \geq \tau_r$}{
      $\mathcal{P}_{\text{set}} \gets \mathcal{P}_{\text{set}} \cup \{\mathcal{P}\}$\;
    }
    
    \If{$\text{used\_options}$ contains all possible options}{
      \textbf{Break}\;
    }
  }
  \KwRet{$\mathcal{P}_{\text{set}}$}\;
\end{algorithm}

\textbf{Translator} is tasked with converting the reasoning steps synthesized by a formal engine into natural language narratives. These narratives serve as high-quality training data to cultivate the reasoning capabilities of Large Language Models (LLMs). As illustrated in ~\cref{fig:translator_illus}, the Translator's process is twofold. First, it employs a state-of-the-art LLM (GPT-4o) to translate each formal reasoning step into its natural language counterpart. Given the highly structured format of the formal steps, we utilize a few-shot prompting strategy. This approach facilitates a step-by-step translation process rather than a single-pass generation of the entire sequence, thereby significantly enhancing translation fidelity.
\label{para:translator}

However, a direct translation, while lexically aligned with human language, still mirrors the discrete, logical leaps of the formal engine, differing substantially from human cognitive processes. A key limitation is the absence of explicit logical bridges between consecutive steps, failing to articulate how one deduction leads to the next. Training an LLM on such data risks encouraging it to overfit to the syntactic structure of the formal expressions, rather than internalizing the underlying reasoning process. To mitigate this, the Translator executes a second crucial task: interleaving bridging rationales. It uses the translated steps as a scaffold and prompts GPT-4o to articulate the intermediate thought process. Specifically, for each step, the model first summarizes the currently established facts, then explicates how these facts logically connect to the subsequent step, and how this progression contributes to reaching the final goal. This method of generating explicit "Connection Thinking" compels the model to engage in a deeper level of inference, ultimately enhancing the reasoning performance of the models trained on this data.

\begin{algorithm}[tb!]
  \SetKwInput{KwRequire}{Require}
  \SetKwInput{KwEnsure}{Ensure}

  \caption{GeoExplore-T}
  \label{algorithm:geo-traceback}

  \KwRequire{Directed graph $\mathcal{G} = (S, S_0, R, \hookrightarrow)$, target statement $s_t \in S$, threshold $\tau_p$}
  \KwEnsure{Trace-back reasoning data $\mathcal{D}$}

  Initialize $\mathcal{D} \gets \emptyset$\;
  Compute all reasoning paths $\mathcal{P}^t_{\text{set}}=\{\mathcal{P}_1^t, \mathcal{P}_2^t, \dots\}$ to $s_t$ using \cref{algorithm:multi-sol}\;
  Randomly sample a statement $s_e \in S$ such that $s_e \notin \text{UpstreamDependencies}(\mathcal{P}^t)$ for any $\mathcal{P}^t \in \mathcal{P}^t_{\text{set}}$\;
  Search for a reasoning path $\mathcal{P}^e$ leading to $s_e$\;
  
  \If {$\exists \mathcal{P}^t \in \mathcal{P}^t_{\text{set}}, \text{Overlap}(\mathcal{P}^e, \mathcal{P}^t) \geq \tau_p$}{
    Add $\mathcal{P}^e \cup \mathcal{P}^t$ to $\mathcal{D}$\;
  }
  
  \KwRet{$\mathcal{D}$}\;
\end{algorithm}

\subsection{Bootstrap Augmentation}
As previously described, TrustGeoGen begins by constructing a complex geometric figure from a completely random initial scenario and then derives reasoning paths for this instance. While this approach is scalable, its efficiency in generating high-quality data is relatively low, as it may randomly produce geometric scenarios that fail to yield meaningful or valuable results. To address this, TrustGeoGen employs an iterative bootstrap strategy to guide the generation process, thereby improving the efficiency of producing high-quality data. In each iteration, TrustGeoGen utilizes the metrics from the Sampler to evaluate the quality of sampled data on a given graph {\small$G$}. If the average quality of the sampled data on a particular graph {\small$G$} is sufficiently high, TrustGeoGen selects this data as the basis for the subsequent iteration’s initial scenario. Specifically, the process can be represented as {\small$P'_{base} = P = \{p_{i+m}^g, p_{j+n}^d\}$}, where {\small$P$} serves as the starting point. The Constructor then incrementally adds new premises to this foundation, increasing the complexity of the scenario. This leads to the generation of {\small$P' = \{p_{i+m+x}^g, p_{j+n+y}^d\}$}. Following this, the Reasoner, Sampler, and Translator collaboratively work to complete graph construction, problem-solving, and translation, ultimately producing a new batch of high-quality data.

\subsection{Multi-solution Data}
\label{sec:multi-solution}

In the field of GPS, exploring multiple solutions to a given problem is a crucial method for understanding geometric relationships in depth. By investigating different solution logics, a model can examine geometric scenarios from various perspectives, achieving a comprehensive understanding of the relationships among geometric elements. TrustGeoGen facilitates the construction of multi-solution datasets for specific geometric scenarios through the graph-building procedure of the modify Reasoner and the solving procedure of the Sampler.

During the graph construction phase, in order to obtain multiple solutions, for each statement {\small$s$}, when the Reasoner repeatedly identifies transitions in the form {\small$S_r \xrightarrow{r} s$}, each identified transition is retained. This retention of multiple transitions provides potential paths for exploring diverse solutions. In the solving process, the Sampler employs an adaptive search algorithm to generate reasoning paths. For any given problem {\small$s_n$}, the Sampler performs multiple path searches. If a statement {\small$s_i$} has multiple outgoing transitions, the Sampler selects a different transition at each iteration for the next reasoning step. Consequently, the reasoning paths obtained during these iterations will differ. It is worth noting that to ensure the generated reasoning paths are acyclic, the Sampler enforces a constraint during path construction. Specifically, when attempting to add a triplet {\small$(S_i, r_s, s)$} to the reasoning path {\small$\mathcal{P}$}, the Sampler checks whether this triplet already exists in {\small$\mathcal{P}$}. If it does, the statement {\small$s$} is not revisited, thereby preventing cyclical paths in the reasoning process. For details of the multi-solution path searching algorithm, \textit{GeoExplore-M}, refer to ~\cref{algorithm:multi-sol}.

\begin{figure}[tb!]
\centering
\includegraphics[width=0.99\linewidth]{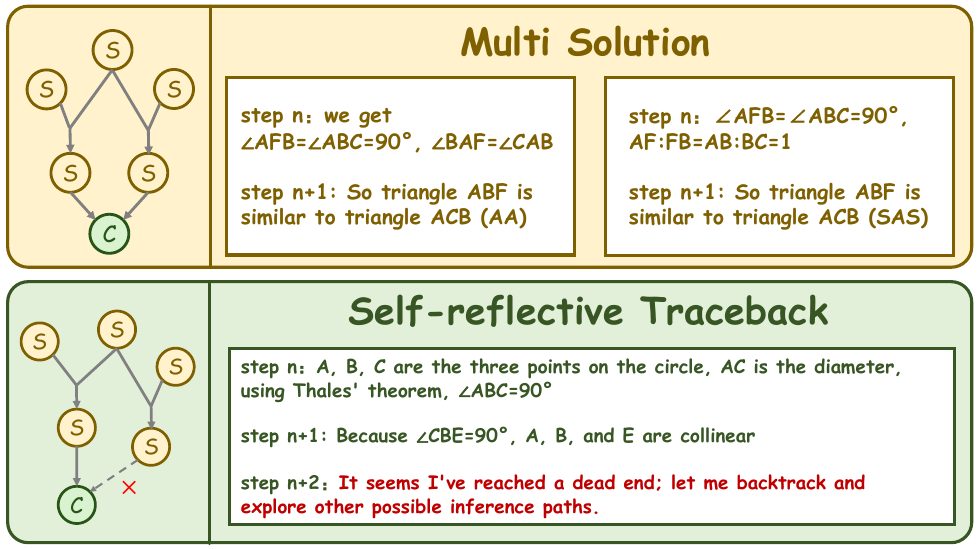}
\caption{Simple examples of Multi-solution and Self-reflective traceback data.}
\label{fig:multi_solution_and_traceback}
\vspace{-5pt}
\end{figure}

\subsection{Self-reflective Traceback Data}
\label{sec:traceback}

\begin{figure*}[tb!]
    \centering
    \begin{subfigure}{0.32\textwidth}
        \includegraphics[width=\linewidth]{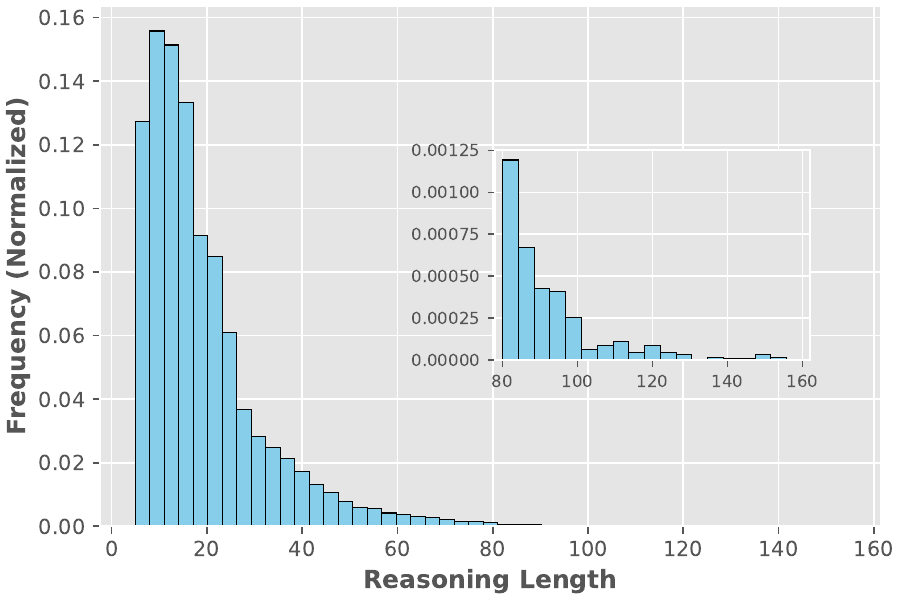}
        \caption{}
        \label{fig:rl_his_all_data}
    \end{subfigure}
    \hfill
    \begin{subfigure}{0.32\textwidth}
        \includegraphics[width=\linewidth]{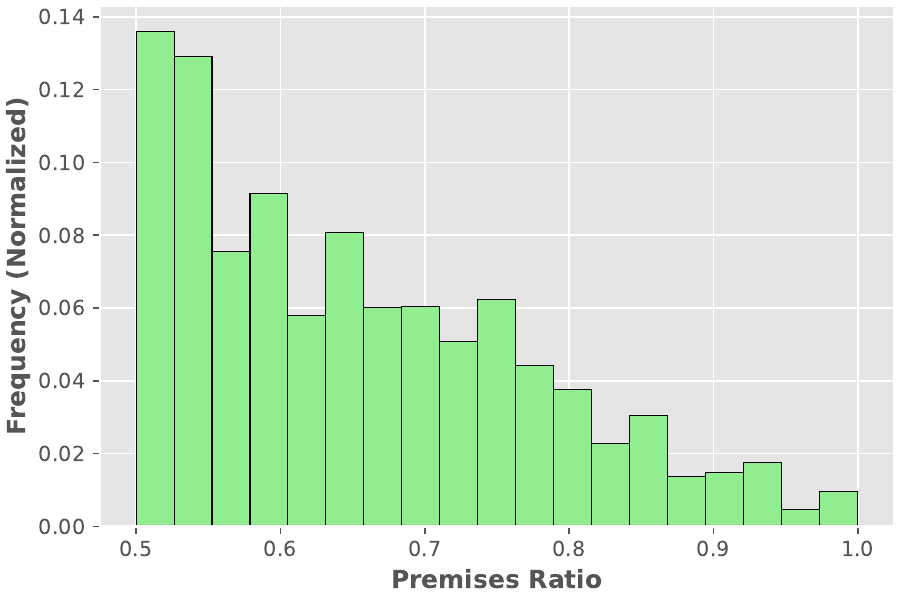}
        \caption{}
        \label{fig:pr_his_all_data}
    \end{subfigure}
    \hfill
    \begin{subfigure}{0.32\textwidth}
        \includegraphics[width=\linewidth]{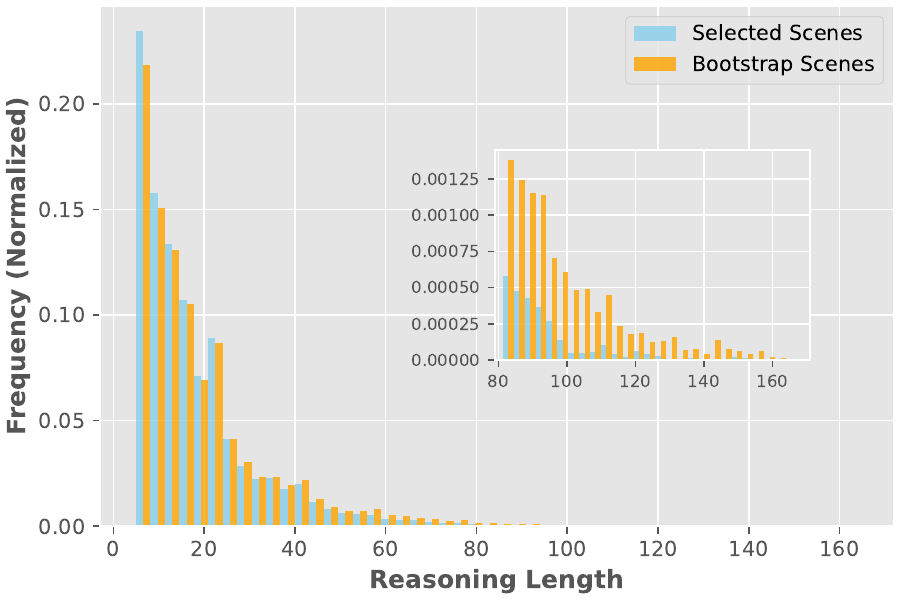}
        \caption{}
        \label{fig:rl_his_bootstrap}
    \end{subfigure}
    \caption{Data distribution and augmentation. (a) Distribution of reasoning lengths, with most samples below 60 steps and a sharp decline beyond 80; the zoomed-in view highlights the ultra-long reasoning range. (b) Distribution of premise ratios, reflecting variability in logical dependencies. (c) Comparison of reasoning lengths before and after bootstrap augmentation, showing increased deep-reasoning samples (reasoning length $\ge$ 40) and reduced shallow ones.}
    \label{fig:data_distribution}
\end{figure*}

\begin{figure*}[tb!]
\centering
\includegraphics[width=0.99\linewidth]{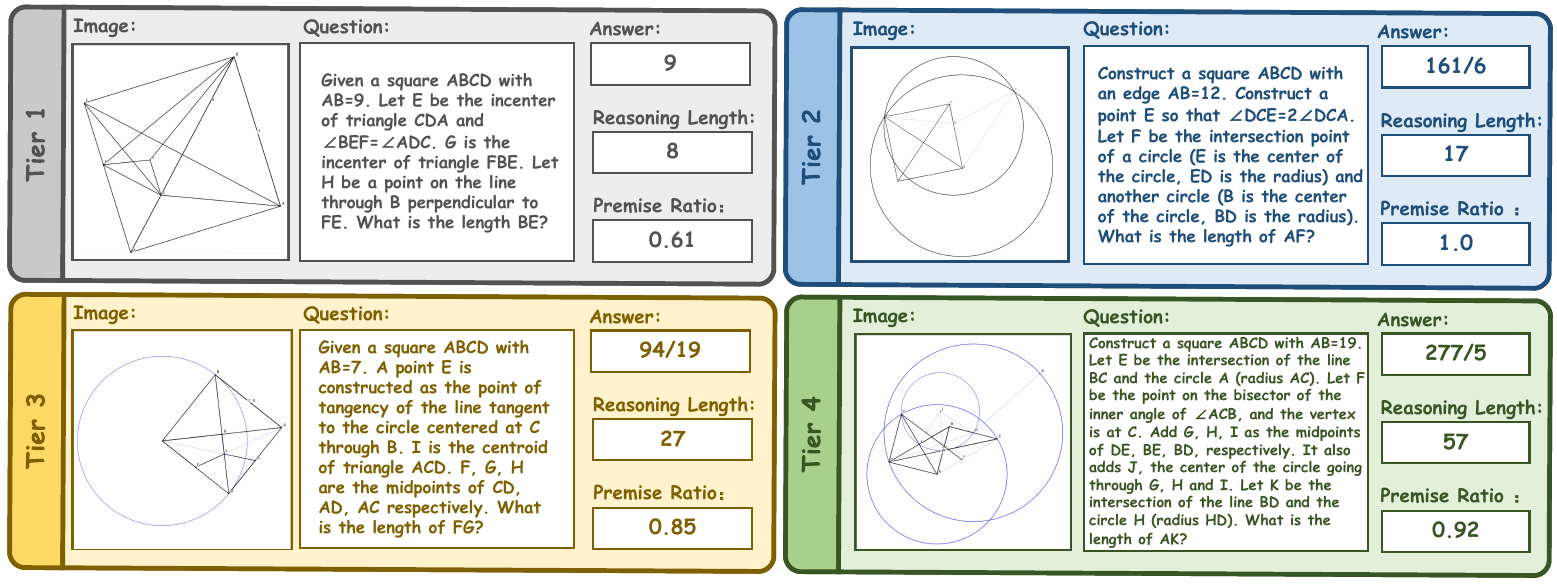}
\vspace{-5pt}
\caption{Visualization examples of different difficulty levels in \textit{GeoTrust-test}, where "reasoning length" indicates step length of reasoning path and "premise ratio" refer to the ratio of premises unutilized during formal reasoning and premises provided in the question.}
\label{fig:testset}
\vspace{-16pt}
\end{figure*}

Recent studies on the reasoning capabilities of large models suggest that these models acquire substantial knowledge during the pre-training phase. By providing training data with various cognitive templates during the post-training phase, the models can better utilize their inherent knowledge, thereby enhancing reasoning ability and improving response quality\citep{ye2025limo, yang2025reasonflux}. Among these cognitive templates, self-reflection trace-back reasoning serve as highly effective approaches. However, self-reflection and retrospective reasoning often occur during the author's cognitive process and are rarely preserved in the final outputs, making the collection of relevant data particularly challenging. These types of data are even more scarce in the domain of multi-modal geometric problem-solving.

TrustGeoGen defines trace-back reasoning data as a sub-graph of {\small$\mathcal{G} = (S, S_0, R, \hookrightarrow)$}, characterized as the union of sub-graphs with identical upstream dependencies but differing final statement. Specifically, for a given piece of trace-back reasoning data, the process involves first deriving an incorrect final statement, then retrospectively navigating a correct reasoning path (shared by both the incorrect and correct paths), and ultimately arriving at the correct final statement. To achieve this, TrustGeoGen selects a target statement {\small$s_t$} and employs~\cref{algorithm:multi-sol} to enumerate all possible reasoning paths {\small$\mathcal{P}^t \in \mathcal{P}^t_{\text{set}}$}. Next, a statement {\small$s_e$} is randomly sampled from the statement set {\small$S$}, ensuring that {\small$s_e$} does not belong to the upstream dependencies of any {\small$\mathcal{P}^t$}. Subsequently, TrustGeoGen searches for a reasoning path {\small$\mathcal{P}^e$} leading to {\small$s_e$}. If {\small$\mathcal{P}^e$} overlaps with any {\small$\mathcal{P}^t \in \mathcal{P}^t_{\text{set}}$} and the overlapping portions exceed the threshold {\small$\tau_p$}, TrustGeoGen outputs {\small$\mathcal{P}^e \cup \mathcal{P}^t$} as trace-back reasoning data. Notably,~\cref{algorithm:multi-sol} is utilized to identify all possible reasoning paths for {\small$s_t$} to ensure that {\small$s_e$} does not appear in any potential upstream dependencies, maintaining the integrity of the reasoning process leading to {\small$s_t$}. The details of the trace-back data acquisition algorithm, \textit{GeoExplore-T}, are presented in ~\cref{algorithm:geo-traceback}.

\section{Data Analysis}
The modality-completed trustworthy geometric reasoning data holds great potential for improving the geometric problem-solving capabilities of MLLMs. Moreover, it could generalize these abilities to other tasks requiring deep reasoning. To validate it, we utilized TrustGeoGen to produce the dataset \textit{GeoTrust} of 200K raw samples\footnote{Raw samples refers to the data which is not processed by translator}, from which 8K were sampled as the training set. Additionally, we employed different thresholds, $\tau_l$ and $\tau_r$, in conjunction with manual screening to curate a testset, GeoTrust-test, with 240 samples of varying levels of difficulty.

\subsection{Data Distribution}

TrustGeoGen was executed on a 60-core Intel Xeon (32) CPU @ 2.900GHz for two days, generating an initial dataset of 200K geometric reasoning instances in formal language. During the data generation process, we set the two filtering thresholds in the Sampler to $\tau_l = 5$ and $\tau_r = 0.5$, respectively. The distributions of the reasoning length and premises ratio for the resulting 200K data are illustrated in ~\cref{fig:rl_his_all_data} and ~\cref{fig:pr_his_all_data}, respectively. The majority of the data is concentrated in the region where the reasoning length is less than 60, and it decreases sharply beyond 80. However, a small portion of the data exhibits reasoning lengths exceeding 150, indicating a considerably high level of reasoning complexity. For a more detailed view of the distribution, please refer to the zoomed-in section of ~\cref{fig:rl_his_all_data}.

\begin{table*}[ht]
	\centering	
    \caption
	{
        Performance of state-of-the-art multi-modal language models on \textit{GeoTrust-test}, which is divided into four difficulty levels. The best results are highlighted in \textbf{bold}, and the second best results are highlighted with an \underline{underline}.
	}
    \vspace{-5pt}
	\resizebox{0.99\textwidth}{!}{
	\begin{tabular}	{l l |  r r  | r r | r  r | r r | r r}
	\toprule	 	
	 \multirow{2}{*}{Model} & \multirow{2}{*}{\#Params} & \multicolumn{2}{c|}{Total (out of 240)}  & \multicolumn{2}{c|}{$Tier_1$ (out of 60)} & \multicolumn{2}{c}{$Tier_2$ (out of 60)} & \multicolumn{2}{c|}{$Tier_3$ (out of 60)} & \multicolumn{2}{c}{$Tier_4$ (out of 60)} \\
     & &\textit{Count} &\textit{Accuracy}&\textit{Count} &\textit{Accuracy}&\textit{Count} &\textit{Accuracy}&\textit{Count} &\textit{Accuracy}&\textit{Count} &\textit{Accuracy} \\
     \midrule
    LLaVA-1.5-7B~\citep{llava/liu2024visual} & 7B & 4 & 1.67\% & 3 & 5.00\% & 1 & 0.42\% & 0 & 0.00\% & 0 & 0.00\%
	\\
    Qwen2-VL-7B~\citep{Qwen2-VL} & 7B & 11 & 4.58\% & 5 & 8.33\% & 2 & 3.33\% & 2 & 3.33\% & 2 & 3.33\% \\ 
    GPT-4o~\cite{openai2024gpt4o}
    & - & 62 & 25.83\% & 31 & 51.67\% & 10 & 16.67\% & 11 & 18.33\% & 10 & 16.67\% \\
    Claude-3.7-sonnet~\citep{2024claude}
    & - & 66 & 27.50\% & 33 & 55.00\% & 16 & 26.67\% & 10 & 16.67\% & 7 & 11.67\% \\
    Qwen2.5-VL-72B~\cite{Qwen2.5-VL}
    & 72B & 68 & 28.33\% & 32 & 53.33\% & 15 & 25.00\% & 15 & 25.00\% & 6 & 10.00\% \\
    DeepSeek-R1 *~\citep{DeepSeekAI2025DeepSeekR1IR}
    & 671B & 89 & 37.08\% & 30 & 50.00\% & 20 & 33.33\% & 22 & 36.67\% & 17 & 28.33\% \\
    Intern-S1~\cite{bai2025intern} & 235B+6B & \underline{104} & \underline{43.33\%} & \underline{35} & \underline{58.33\%} & \underline{21} & \underline{35.00\%} & \underline{25} & \underline{41.67\%} & \textbf{23} & \textbf{38.33\%} \\
    Gemini-2.5-pro~\cite{comanici2025gemini}
    & - & \underline{104} & \underline{43.33\%} & 34 & 56.67\% & \textbf{24} & \textbf{40.00\%} & \textbf{26} & \textbf{43.33\%} & \underline{20} & \underline{33.33\%} \\
    OpenAI-o3~\cite{openai_o3}
    & - & \textbf{110} & \textbf{45.83\%} & \textbf{37} & \textbf{61.67\%} & \textbf{24} & \textbf{40.00\%} & \textbf{26} & \textbf{43.33\%} & \textbf{23} & \textbf{38.33\%} \\
    \bottomrule
	\end{tabular}
	}
    \vspace{0.3em}
    \par 
    \raggedright 
    \footnotesize 
    \hspace{1em}* For DeepSeek-R1, we only provide natural language questions, but no images.\\
	\label{tab:testset_sota}
\end{table*}

\subsection{Deep Reasoning Augmentation}

As mentioned earlier, the randomly constructed scenarios by TrustGeoGen exhibit a rapid decline in reasoning length within the ultra-long range, posing challenges for generating deep-reasoning problems. To address this, we selected 226 samples with the longest reasoning lengths from an initial dataset of 200K and applied bootstrap augmentation to increase the proportion of deep-reasoning problems. Finally we obtained 376 geometric scenes after applying the bootstrap method. As shown in ~\cref{fig:rl_his_bootstrap}, after the bootstrap process, the data distribution in the region where reasoning length is smaller than 40 decreased, while it increased in the region where reasoning length is larger than 40. This method can be repeatedly applied to efficiently construct reasoning problems with significant depth and complexity.

\subsection{Construction and Analysis of GeoTrust-Test}
\label{sec:testset}
The synthetic data generated by TrustGeoGen can also be utilized to evaluate a model's capabilities in the domain of deep geometric reasoning. From an initial dataset of 200K samples, we manually curated a set of 240 high-quality problems as the test set, graded by different levels of difficulty. Unlike the training data, which incorporates a mix of proof and solution-type problems, the \textit{GeoTrus-test} is composed exclusively of problems with numerical solutions to simplify the evaluation process. As shown in ~\cref{fig:testset}, these problems are divided into four tiers, with each tier containing 60 problems. The reasoning lengths for $Tier_1$ range from 5 to 10 steps, $Tier_2$ spans from 10 to 20 steps, $Tier_3$ ranges from 20 to 50 steps, and $Tier_4$ exceeds 50 steps. It is worth noting that, to introduce distractors into the questions, the premises provided may not necessarily all be utilized in the reasoning process, resulting in a potentially lower premise ratio.
\section{Experiments}

To validate the quality and effectiveness of the geometric Problem Solving data generated by TrustGeoGen, we conducted experiments to explore:
\begin{itemize}
\vspace{0.3em}
\item  \textit{Whether existing MLLMs demonstrate competent performance on complex geometric problems? (\cref{sec:existing_performance})}
\vspace{0.3em}
\item  \textit{Whether the data generated by TrustGeoGen have sufficient complexity to provide gains for SOTA models? (\cref{sec:complexity_analysis})}
\vspace{0.3em}
\item \textit{Whether geometric reasoning data with rigorous formal verification improve MLLMs' capabilities and provide specific advantages? (\cref{sec:effectiveness_analysis})}
\vspace{0.3em}
\item \textit{Whether \textit{GeoTrust}, constructed without prior data sources, can generalize to OOD geometric testset? (\cref{sec:ood_generalization})}
\end{itemize}

\begin{table*}[htbp]
	\centering	
    \caption
	{
        Complexity comparison between different benchmarks. The most difficult results are highlighted in \textbf{bold}, and the second difficult results are highlighted with an \underline{underline}. It is indicated that our geometry constructed problems are more challenging for most mainstream MLLMs.
	}
    \vspace{-5pt}
	\resizebox{\textwidth}{!}{ 
	\begin{tabular}	{l c | c c c c}
	\toprule	 	
	\textbf{Model} & \textbf{Release Date} & \makecell{\textbf{GeoQA} \\ (mid-school level)} & \makecell{\textbf{Geometry3K} \\ (mid-school level)} & \makecell{\textbf{OlympiadBench-Geo} \\ (olympiad level)} & \textbf{GeoTrust-test} \\
     \midrule
    GPT-4o~\cite{openai2024gpt4o} & May, 2024 & 42.31\% & 31.45\% & \textbf{13.39\%} & \underline{25.83\%} \\ 
    Claude-3.7-sonnet~\citep{2024claude} & Feb, 2025 & 49.73\% & 33.28\% & \textbf{17.86\%} & \underline{27.50\%} \\
    Qwen2.5-VL-72B~\cite{Qwen2.5-VL} & Jan, 2025 & 67.90\% & 35.44\% & \underline{29.46\%} & \textbf{28.33\%} \\ 
    Intern-S1~\cite{bai2025intern} & Jul, 2025 & 62.47\% & 52.31\% & \underline{49.11\%} & \textbf{43.33\%} \\
    Gemini-2.5-pro\cite{comanici2025gemini} & Jun, 2025 & 79.58\% & 80.70\% & \underline{75.00\%} & \textbf{43.33\%} \\
    OpenAI-o3~\cite{openai_o3} & Apr, 2025 & 79.31\% & 81.03\% & \underline{77.68\%} & \textbf{45.83\%} \\ 
    \bottomrule
	\end{tabular}
	}
	\label{tab:bench_compare} 
    \vspace{-5pt}
\end{table*}

\subsection{Dataset, Metric, and Implementation Detail}
\label{sec:exp_details}

\textbf{Dataset.} To evaluate the capability of MLLMs in solving complex geometry problems, we construct \textit{GeoTrust-test} by partitioning the original \textit{GeoTrust} dataset. As introduced in~\cref{sec:testset}, the testset \textit{GeoTrust-test} comprehensively covers four difficulty levels ranging from $Tier_1$ to $Tier_4$. To validate the effectiveness of our data construction, we additionally sample 2342 geometry problems as \textit{GeoTrust-train}, with different difficulty tier.

\noindent \textbf{Metric.} To eliminate potential evaluation bias introduced by the multiple-choice format, all input questions in the experiments are presented without answer options. Both the model's final output answers and GT labels are extracted and converted into floating-point values for numerical comparison, with a relative error tolerance of 1\% permitted.

\noindent \textbf{Implementation Detail.} In the experimental training protocol, all models are trained through supervised fine-tuning (SFT) with full-parameter updates for one epoch to prevent overfitting. All training and evaluation processes are conducted on 8 NVIDIA A100 (80G) GPUs.

\subsection{Performance of Existing MLLMs}
\label{sec:existing_performance}

As shown in~\cref{tab:testset_sota}, the experimental results from the \textit{GeoTrust-test} testset reveal critical insights into the capabilities of MLLMs in tackling complex geometric problems. Existing open-source models, such as Qwen2-VL-7B, demonstrate significant limitations, achieving only a 4.58\% overall accuracy (11/240), which also underscores the inherent challenges of the dataset and confirms its independence from prior biases in existing open-source testsets. The closed-source reasoning model, OpenAI-o3, demonstrated the strongest geometric reasoning capability, solving 110 problems with an accuracy of 45.83\%. Furthermore, models that have undergone "deep thinking" training (e.g., OpenAI-o3, Gemini-2.5-pro, Intern-S1, DeepSeek-R1) exhibited a more gradual decline in accuracy as problem difficulty increased (from \textit{Tier1} to \textit{Tier4}). In contrast, models without such specialized training (e.g., GPT-4o, Claude-3.7-sonnet, Qwen2.5-VL-72B) experienced a sharp drop in accuracy on more challenging problems. This distinction underscores the critical importance of deep reasoning abilities for GPS. The consistent decrease in accuracy across all models with rising difficulty collectively emphasizes the need for enhanced reasoning architectures and training paradigms to address the steep complexity gradient in geometric problem solving.

\subsection{Complexity Analysis Between Benchmarks}
\label{sec:complexity_analysis}

Although constructing geometric data using formal engines can enhance acquisition efficiency, it presents challenges in ascertaining the data's complexity and its actual contribution to model training. To address this, TrustGeoGen utilizes filtering parameters, $\tau_l$ and $\tau_r$, to screen for problems of varying difficulty, thereby enabling the selective generation of data that provides tangible benefits to the model. As detailed in~\cref{tab:bench_compare}, we evaluated several state-of-the-art (SOTA) multimodal large language models on four distinct geometry benchmarks to assess their respective complexity. The performance of SOTA models indicates that \textit{GeoTrust-test} presents a level of difficulty comparable to OlympiadBench-Geo, with models generally struggling to achieve high scores on either. Notably, even well-pretrained models like Gemini-2.5-pro and OpenAI-o3, which exhibit substantially improved accuracy on OlympiadBench-Geo thanks to extensive pre-training on geometry data, still fail to surpass the 50\% accuracy mark on \textit{GeoTrust-test}. This underscores the potential of \textbf{TrustGeoGen} to further boost the performance of these advanced models by providing a continuous stream of fresh and challenging data.

\subsection{Effectiveness Analysis of Trustworthy Data}
\label{sec:effectiveness_analysis}

TrustGeoGen constructs trustworthy geometric reasoning data using a formal engine. It bridges the gap between formal and human-like reasoning through "connection thinking" (\cref{para:translator}). Furthermore, thinking template data—comprising multi-solution data (\cref{sec:multi-solution}) and self-reflective traceback data (\cref{sec:traceback})—is utilized to introduce diverse reasoning pathways. In this section, we design comprehensive experiments to validate the effectiveness of our trustworthy data.

\begin{table*}
	\centering	
    \caption
	{
        Translation performance comparison across different models and training data sources. The $\Delta$ column indicates the improvement over the "Pretraining" baseline. Improvements are marked with an upward arrow ($\uparrow$).
	}
    \vspace{-5pt}
	\resizebox{\textwidth}{!}{
	\begin{tabular}	{l | c r | c r | c r | c r}
	\toprule	 	
	\multirow{2}{*}{\textbf{Training Data}} & \multicolumn{2}{c|}{\textbf{LLaVA-1.5-7B}} & \multicolumn{2}{c|}{\textbf{LLaVA-1.5-13B}} & \multicolumn{2}{c|}{\textbf{Qwen2-VL-2B}} & \multicolumn{2}{c}{\textbf{Qwen2-VL-7B}} \\
	& Accuracy & \multicolumn{1}{c|}{$\Delta$} & Accuracy & \multicolumn{1}{c|}{$\Delta$} & Accuracy & \multicolumn{1}{c|}{$\Delta$} & Accuracy & \multicolumn{1}{c}{$\Delta$} \\
    \midrule
    Pretraining & 1.67\% (4) & - & 2.08\% (5) & - & 3.33\% (8) & - & 4.58\% (11) & - \\
    NL Translation & 6.25\% (15) & 4.58\%$\uparrow$ & 6.67\% (16) & 4.59\%$\uparrow$ & 8.75\% (21) & 5.42\%$\uparrow$ & 8.75\% (21) & 4.17\%$\uparrow$ \\
    Connection Thinking & 15.42\% (37) & 13.75\%$\uparrow$ & 19.17\% (46) & 17.09\%$\uparrow$ & 13.75\% (33) & 10.42\%$\uparrow$ & 21.67\% (52) & 17.09\%$\uparrow$ \\
    \bottomrule
	\end{tabular}
	}
	\label{tab:ct_necessary}
    \vspace{-0.2em}
\end{table*}

\subsubsection{Necessary of Connection Thinking}

We conducted experiments to validate our data's effectiveness using four models: LLaVA-1.5-7B, LLaVA-1.5-13B, Qwen2-VL-2B, and Qwen2-VL-7B. Each model was fine-tuned via Supervised Fine-Tuning (SFT) on a sampled dataset of 2,158 instances. We applied two distinct training strategies separately: one using the NL translation data and the other using the connection thinking data, both of which are detailed in~\cref{fig:translator_illus}. All fine-tuned models were subsequently evaluated on the \textit{GeoTrust-test} benchmark. 

As shown in~\cref{tab:ct_necessary}, the results demonstrate that both training approaches lead to a significant increase in model accuracy. Notably, the models trained with the connection thinking data exhibited a more substantial improvement, consistently achieving accuracy gains of over 10\%.

\begin{figure}[tbp!]
\centering
\includegraphics[width=0.90\linewidth]{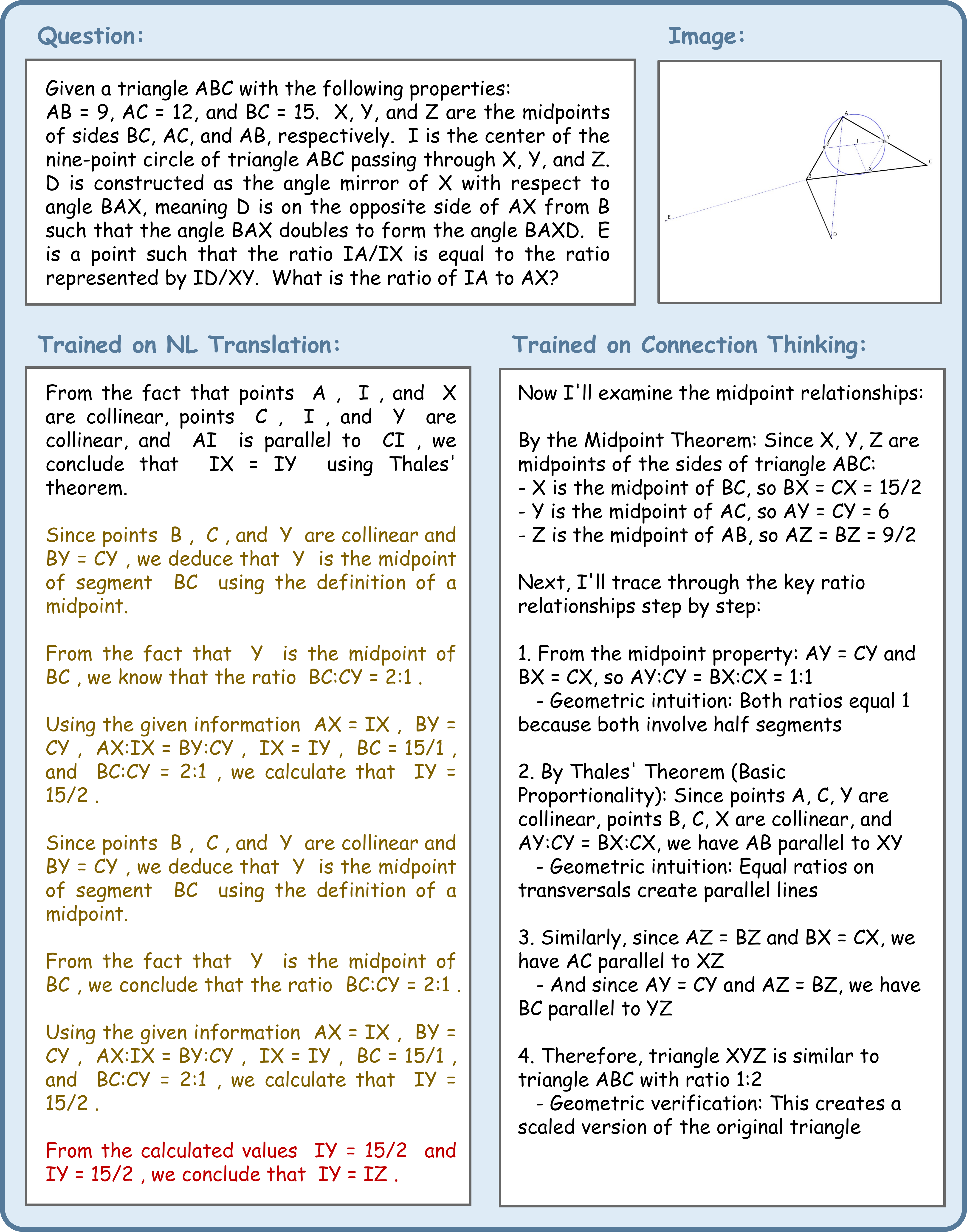}
\caption{Fragments of model’s response: The response of model trained on NL Translation dataset is \textbf{repeating} \& \textbf{illogical}, while the response of model trained on Connection Thinking dataset is reasonable and clear.}
\label{fig:ct_necessary}
\vspace{-5pt}
\end{figure}

Furthermore, we analyzed the outputs of models trained on different datasets. As illustrated in~\cref{fig:ct_necessary}, there is a substantial difference in the content generated by models trained on the two respective datasets. The model trained on the "NL Translation" data tends to produce repetitive content and exhibits apparent logical fallacies. In contrast, the model trained on our proposed "connection thinking" data generates outputs that are significantly more coherent and logically sound.

The "NL Translation" data suffers from two key limitations that prevent a model from truly acquiring reasoning capabilities. First, its formal-language reasoning steps are highly structured, merely presenting premises, applied theorems, and derived conclusions in a templated format without elucidating the underlying rationale of the proof process. A model trained on such data learns to replicate theorems it has seen rather than how to apply them proficiently. Second, a significant gap exists between this formal reasoning process and authentic human-like reasoning, as there are no explicit logical connections between adjacent steps. Consequently, each subsequent step appears to emerge abruptly, lacking logical coherence.

To address these limitations, our "connection thinking" approach uses the "NL Translation" data as a foundational skeleton but augments it by inserting a detailed thought process between each pair of consecutive reasoning steps. This enrichment not only explains each step in detail but also articulates the logical link between them by considering previously established conclusions and the ultimate goal. This enriched data structure enables the model to master the application of plane geometry theorems, thereby significantly improving its geometric problem-solving capabilities.

\subsubsection{Advantages Provided by Trustworthy Data}

To validate the effectiveness of data generated by TrustGeoGen, we conducted a comparative experiment using pseudo-labeled data. We fed 2,158 questions generated by TrustGeoGen into OpenAI-o3~\cite{openai_o3} and obtained answers from the model. The responses were then evaluated using the metrics described in~\cref{sec:exp_details}, resulting in 846 correctly answered questions, which we treat as pseudo-labeled data. For fairness, we also used the same 846 questions as the GeoTrust Data in our experiments.

We trained four models using the pseudo-labeled data and GeoTrust data with varying proportions (100\%, 50\%, 30\%, and 10\%). The trained models were subsequently tested on the GeoTrust-test set. As shown in~\cref{tab:pseudo_data}, the model trained with 100\% GeoTrust data slightly outperformed the model trained with pseudo-labeled data. This indicates that, given the same data volume, the synthetic data generated by TrustGeoGen achieves comparable or even superior results compared to pseudo-labeled data produced by SOTA models. It is worth noting that the SOTA model did not correctly answer all TrustGeoGen-generated questions, whereas TrustGeoGen can generate large volumes of data accompanied by accurate solution processes, providing a significant advantage for model training. Furthermore, the experimental results demonstrate that the model’s performance consistently improves with increasing training data size, further confirming the effectiveness of our generated data.

\begin{table*}
	\centering	
    \caption
	{
        Performance comparison using different data augmentation strategies. "Pseudo-labels" and various percentages of "Our data" are compared against the "Pretraining" baseline. The $\Delta$ column shows the improvement over the baseline. Improvements are marked with an upward arrow ($\uparrow$).
	}
    \vspace{-5pt}
	\resizebox{\textwidth}{!}{
	\begin{tabular}	{l | c r | c r | c r | c r}
	\toprule	 	
	\multirow{2}{*}{\textbf{Training data}} & \multicolumn{2}{c|}{\textbf{LLaVA-1.5-7B}} & \multicolumn{2}{c|}{\textbf{LLaVA-1.5-13B}} & \multicolumn{2}{c|}{\textbf{Qwen2-VL-2B}} & \multicolumn{2}{c}{\textbf{Qwen2-VL-7B}} \\
	& Accuracy & \multicolumn{1}{c|}{$\Delta$} & Accuracy & \multicolumn{1}{c|}{$\Delta$} & Accuracy & \multicolumn{1}{c|}{$\Delta$} & Accuracy & \multicolumn{1}{c}{$\Delta$} \\
    \midrule
    Pretraining & $1.67\% (4)$ & - & $2.08\% (5)$ & - & $3.33\% (8)$ & - & $4.58\% (11)$ & - \\
    Pseudo-labels data & $12.08\% (29)$ & $10.42\%\uparrow$ & $13.75\% (33)$ & $11.67\%\uparrow$ & $11.25\% (27)$ & $7.92\%\uparrow$ & $12.92\% (31)$ & $8.33\%\uparrow$ \\
    GeoTrust data $10\%$ & $5.83\% (14)$ & $4.17\%\uparrow$ & $7.50\% (18)$ & $5.42\%\uparrow$ & $3.75\% (9)$ & $0.42\%\uparrow$ & $11.25\% (27)$ & $6.67\%\uparrow$ \\
    GeoTrust data $30\%$ & $9.17\% (22)$ & $7.50\%\uparrow$ & $10.83\% (26)$ & $8.75\%\uparrow$ & $10.42\% (25)$ & $7.08\%\uparrow$ & $12.08\% (29)$ & $7.50\%\uparrow$ \\
    GeoTrust data $50\%$ & $9.58\% (23)$ & $7.92\%\uparrow$ & $11.25\% (27)$ & $9.17\%\uparrow$ & $10.00\% (24)$ & $6.67\%\uparrow$ & $12.50\% (30)$ & $7.92\%\uparrow$ \\
    GeoTrust data $100\%$ & $12.92\% (31)$ & $11.25\%\uparrow$ & $14.17\% (34)$ & $12.08\%\uparrow$ & $12.50\% (30)$ & $9.17\%\uparrow$ & $14.17\% (34)$ & $9.58\%\uparrow$ \\
    \bottomrule
	\end{tabular}
	}
	\label{tab:pseudo_data}
    \vspace{-0.2em}
\end{table*}

\subsubsection{Advantages of by Thinking Template}
\label{sec:ablation_of_tt}

We investigated the impact of our proposed thinking template across four model architectures. Our primary training dataset, \textit{GeoTrust-train}, was constructed by augmenting a baseline dataset of 2,158 instances with 184 instances that incorporate the thinking template, resulting in a total of 2,342 instances. To isolate the contribution of the Thinking Template, we created three training setups:

\begin{itemize}
    \item \textit{GeoTrust-train} (w/o TT): This version consists solely of the 2,158 baseline instances, with all Thinking Template data removed.
    \item \textit{GeoTrust-train} (Re TT): To maintain the dataset size for a fair comparison, this version replaces the 184 Thinking Template instances with 184 additional baseline instances, also totaling 2,342 instances.
    \item \textit{GeoTrust-train}: This is the full dataset including both the 2,158 baseline instances and the 184 Thinking Template instances.
\end{itemize}

Subsequently, each of the three resulting models was evaluated on the \textit{GeoTrust-test} to assess their performance.

\begin{figure}[t]
\centering
\includegraphics[width=0.99\linewidth]{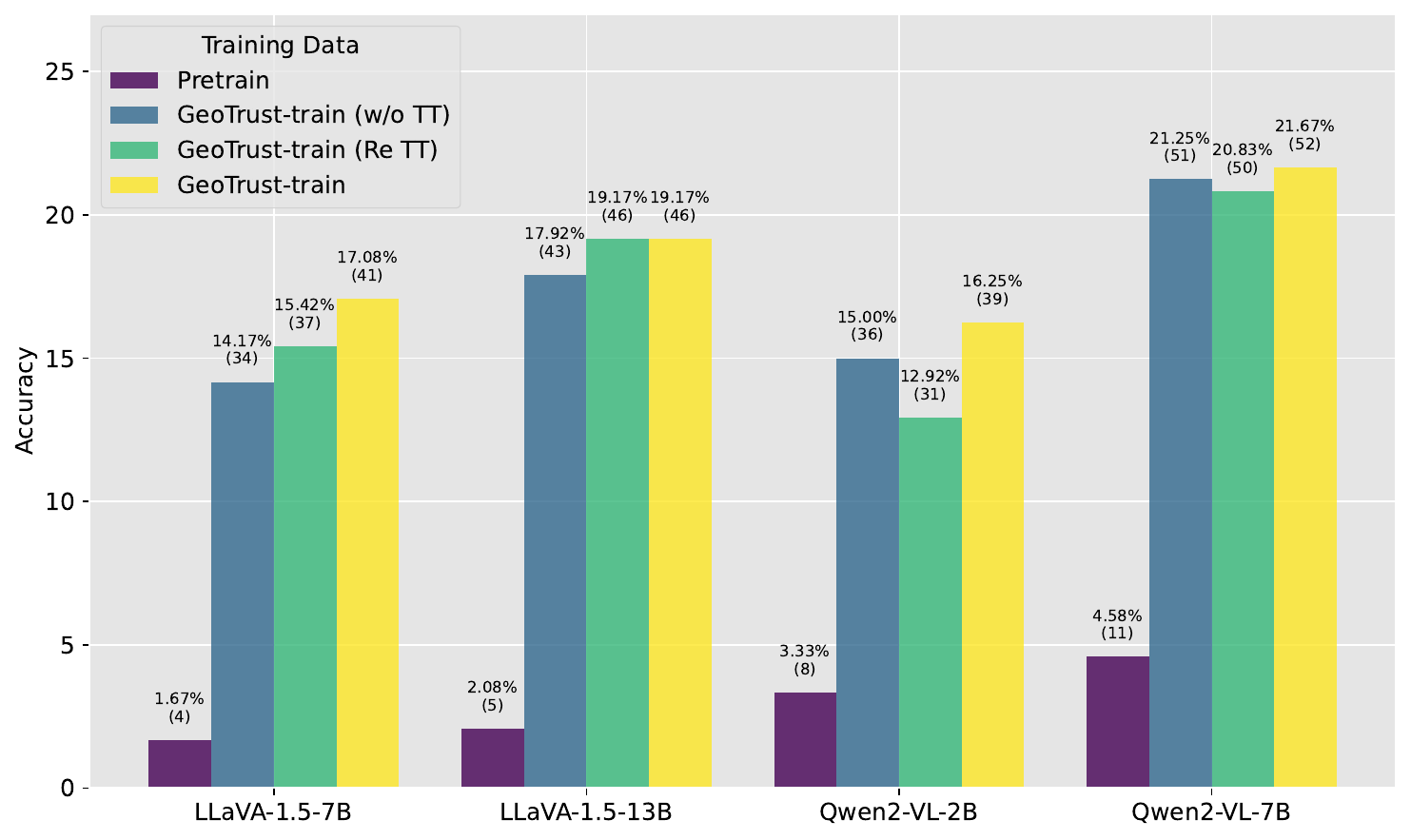}
\vspace{-6pt}
\caption{Ablation Study of Thinking Template on \textit{GeoTrust-test}}
\label{fig:tt_ablation_in_domain}
\vspace{-5pt}
\end{figure}

Fig.~\ref{fig:tt_ablation_in_domain} illustrates the experimental results on the Thinking Template dataset. The Thinking Template demonstrates enhanced performance across all four models. Interestingly, for the LLaVA-1.5 series, the model trained with our Thinking Template (Re TT) consistently outperformed the version trained without it (w/o TT). In contrast, for the Qwen2-VL series, the model trained without the Thinking Template (w/o TT) achieved superior results.

\subsection{OOD Generalization}
\label{sec:ood_generalization}

We evaluated the efficacy of TrustGeoGen in enhancing GPS performance by conducting extensive experiments across three diverse out-of-domain (OOD) benchmarks: GeoQA, Geometry3k, and OlympiadBench. These datasets were strategically chosen to cover a wide spectrum of challenges. GeoQA, sourced from real-world scenarios, offers diverse and unstructured problems accompanied by rich natural language rationales. Geometry3k provides formally structured problems generated via a formal engine, but without natural language explanations. To assess performance on highly complex tasks, we included OlympiadBench, which contains challenging competition-level problems. Together, this suite of benchmarks enables a robust and comprehensive evaluation of our approach.

\subsubsection{GeoQA}

The GeoQA dataset originates from real-world sources and comprises a high-quality, diverse set of problems presented in natural language along with their corresponding solutions. The dataset provides a training partition of 3,499 problems and a test partition of 754. In our experiments, we fine-tuned four distinct models via Supervised Fine-Tuning (SFT) utilizing three separate training datasets: the GeoQA training data, the \textit{GeoTrust-train} data, and a combined dataset. Subsequently, we assessed the models' efficacy on the GeoQA test set.

The results, presented in~\cref{tab:geoqa_think_data}, indicate that for the LLaVA-1.5 model family, all three training approaches yielded performance enhancements. It is particularly noteworthy that \textit{GeoTrust-train}, despite being from a different domain, still enabled the models to achieve superior results on GeoQA. Moreover, a synergistic effect was observed when the two datasets were merged, with \textit{GeoTrust-train} offering further incremental gains.

In contrast, for the Qwen2-VL model family, fine-tuning exclusively on the \textit{GeoTrust-train} data led to performance degradation. A plausible explanation is that the Qwen2-VL series, being more recent, already possesses strong geometric reasoning abilities from extensive pre-training. This strong prior knowledge explains its superior initial performance, but also makes it susceptible to performance loss when adapted to a dataset with a distinct data distribution like \textit{GeoTrust-train}. Despite this, using \textit{GeoTrust-train} as a supplementary dataset for GeoQA still enhanced the models' performance.

\begin{table}[tbp]
\vspace{-0.3em}
	\centering	
    \caption
	{
        Performance comparison on GeoQA dataset using different training data mixture. The $\Delta$ column indicates the improvement over the baseline. An up-arrow ($\uparrow$) indicates an improvement, and a down-arrow ($\downarrow$) indicates a decline.
	}
    \vspace{-5pt}
	\resizebox{0.49\textwidth}{!}{
	\begin{tabular}	{l | l | c | r} 
	\toprule	 	
	\textbf{Model} & \textbf{Training Data} & \textbf{Accuracy} & \multicolumn{1}{c}{\textbf{$\Delta$}} \\
     \midrule
    \multirow{4}{*}{LLaVA-1.5-7B} & - & 13.66\% (103) & - \\  
    & GeoQA & 24.80\% (187) & $+11.14\%\,\uparrow$ \\
    & GeoTrust-train & 13.79\% (104) & $+0.13\%\,\uparrow$ \\
    & GeoQA+GeoTrust-train & 26.92\% (203) & $+13.26\%\,\uparrow$ \\  
    \midrule
    \multirow{4}{*}{LLaVA-1.5-13B} & - & 16.31\% (123) & - \\  
    & GeoQA & 25.46\% (192) & $+9.15\%\,\uparrow$ \\
    & GeoTrust-train & 16.71\% (126) & $+4.00\%\,\uparrow$ \\
    & GeoQA+GeoTrust-train & 30.11\% (227) & $+13.80\%\,\uparrow$ \\  
    \midrule
    \multirow{4}{*}{Qwen2-VL-2B} & - & 20.56\% (155) & - \\  
    & GeoQA & 25.46\% (192) & $+4.90\%\,\uparrow$ \\
    & GeoTrust-train & 14.99\% (113) & $-5.57\%\,\downarrow$ \\
    & GeoQA+GeoTrust-train & 27.72\% (209) & $+7.16\%\,\uparrow$ \\  
    \midrule
    \multirow{4}{*}{Qwen2-VL-7B} & - & 37.00\% (279) & - \\  
    & GeoQA & 39.66\% (299) & $+2.66\%\,\uparrow$ \\
    & GeoTrust-train & 28.38\% (214) & $-8.62\%\,\downarrow$ \\
    & GeoQA+GeoTrust-train & 43.50\% (328) & $+6.50\%\,\uparrow$ \\  
    \bottomrule
	\end{tabular}
	}
    \par 
    \vspace{0.3em}
    \raggedright 
    \footnotesize 
    \hspace{1em} GeoQA was evaluated in a direct question-answering (QA) format, rather than a multiple-choice setting. The same metrics as described in~\cref{sec:exp_details} were used.
	\label{tab:geoqa_think_data}
    \vspace{-0.2em}
\end{table}

\subsubsection{Geometry3K}

Geometry3K, a dataset constructed via a formal engine for solving geometry problems with formal language, comprises 2,101 training and 601 test examples. We performed SFT experiments on this data using different mixtures, and the results are summarized in~\cref{tab:geometry3k_think_data}. For the LLaVA-1.5-7B, LLaVA-1.5-13B, and Qwen2-VL-2B models, every data mixture enhanced performance. Conversely, on Qwen2-VL-7B, training solely on Geometry3K led to a decline in performance. This performance drop persisted even when the training data was supplemented with \textit{GeoTrust-train}.

Qwen2-VL-7B's strong pre-trained performance on Geometry3K is characteristic of a well-saturated large multimodal model. This suggests that for such models, fine-tuning on data with a similar distribution to the pre-training set can paradoxically lead to performance degradation. Conversely, the smaller Qwen2-VL-2B model, whose pre-training was likely less saturated, still demonstrated further gains when fine-tuned on this data.

\begin{table}[tbp]
\vspace{-0.3em}
	\centering	
    \caption
	{
        Performance comparison on Geometry3k using different training data mixture. 
	}
    \vspace{-5pt}
	\resizebox{0.49\textwidth}{!}{
	\begin{tabular}	{l | l | c | r} 
	\toprule	 	
	\textbf{Model} & \textbf{Training Data} & \textbf{Accuracy} & \multicolumn{1}{c}{\textbf{$\Delta$}} \\
     \midrule
    \multirow{4}{*}{LLaVA-1.5-7B} & - & 1.16\% (7) & - \\  
    & Geometry3K & 6.66\% (40) & $+5.50\%\,\uparrow$ \\
    & GeoTrust-train & 2.83\% (17) & $+1.67\%\,\uparrow$ \\
    & Geometry3K+GeoTrust-train & 7.99\% (48) & $+6.83\%\,\uparrow$ \\  
    \midrule
    \multirow{4}{*}{LLaVA-1.5-13B} & - & 2.33\% (14) & - \\  
    & Geometry3K & 7.32\% (44) & $+4.99\%\,\uparrow$ \\
    & GeoTrust-train & 3.33\% (20) & $+1.00\%\,\uparrow$ \\
    & Geometry3K+GeoTrust-train & 8.65\% (52) & $+6.32\%\,\uparrow$ \\  
    \midrule
    \multirow{4}{*}{Qwen2-VL-2B} & - & 8.32\% (50) & - \\  
    & Geometry3K & 10.15\% (61) & $+1.83\%\,\uparrow$ \\
    & GeoTrust-train & 8.65\% (52) & $+0.33\%\,\uparrow$ \\
    & Geometry3K+GeoTrust-train & 12.15\% (73) & $+3.83\%\,\uparrow$ \\  
    \midrule
    \multirow{4}{*}{Qwen2-VL-7B} & - & 17.97\% (108) & - \\  
    & Geometry3K & 15.80\% (95) & $-2.17\%\,\downarrow$ \\
    & GeoTrust-train & 19.13\% (115) & $+1.16\%\,\uparrow$ \\
    & Geometry3K+GeoTrust-train & 16.14\% (97) & $-1.83\%\,\downarrow$ \\
    \bottomrule
	\end{tabular}
	}
    \par 
    \vspace{0.3em}
    \raggedright 
    \footnotesize 
    \hspace{1em} Geometry3K was evaluated in a direct QA format, rather than a multiple-choice setting. The same metrics as described in~\cref{sec:exp_details} were used.
	\label{tab:geometry3k_think_data}
    \vspace{-0.2em}
\end{table}

\subsubsection{OlympiadBench-geo}

OlympiadBench serves as an Olympiad-level benchmark comprising authentic competition problems to assess the multimodal reasoning capabilities of foundation models. For our evaluation, we designated a subset of 112 geometry problems from OlympiadBench as the test set OlympiadBench-geo to verify the efficacy of our proposed data. In the absence of a provided training set from OlympiadBench, we formulated several data compositions for our experiments by utilizing the training data from GeoQA.

The experimental results, presented in ~\cref{tab:olympiadbench_think_data}, reveal a clear trend. Fine-tuning the four models exclusively on \textit{GeoTrust-train} resulted in universal performance enhancements. However, for the Qwen2-VL-7B model, reliance on GeoQA data alone caused a decline in performance—a phenomenon likely associated with data distribution issues with its pre-training data, as discussed in previous sections. Crucially, combining the GeoQA and \textit{GeoTrust-train} datasets produced marked performance improvements across all models, demonstrating their enhanced problem-solving abilities on this demanding benchmark.

\begin{table}[tbp]
\vspace{-0.3em}
	\centering	
    \caption
	{
        Performance comparison on OlympiadBench-geo using different training data mixture.
	}
    \vspace{-5pt}
	\resizebox{0.49\textwidth}{!}{
	\begin{tabular}	{l | l | c | r} 
	\toprule	 	
	\textbf{Model} & \textbf{Training Data} & \textbf{Accuracy} & \multicolumn{1}{c}{\textbf{$\Delta$}} \\
     \midrule
    \multirow{4}{*}{LLaVA-1.5-7B} & - & 0.00\% (0) & - \\  
    & GeoQA & 3.57\% (4) & $+3.57\%\,\uparrow$ \\
    & GeoTrust-train & 1.79\% (2) & $+1.79\%\,\uparrow$ \\
    & GeoQA+GeoTrust-train & 3.57\% (4) & $+3.57\%\,\uparrow$ \\  
    \midrule
    \multirow{4}{*}{LLaVA-1.5-13B} & - & 1.79\% (2) & - \\  
    & GeoQA & 6.25\% (7) & $+4.46\%\,\uparrow$ \\
    & GeoTrust-train & 2.68\% (3) & $+0.89\%\,\uparrow$ \\ 
    & GeoQA+GeoTrust-train & 6.25\% (7) & $+4.46\%\,\uparrow$ \\  
    \midrule
    \multirow{4}{*}{Qwen2-VL-2B} & - & 2.68\% (3) & - \\  
    & GeoQA & 4.46\% (5) & $+1.78\%\,\uparrow$ \\
    & GeoTrust-train & 4.46\% (5) & $+1.78\%\,\uparrow$ \\
    & GeoQA+GeoTrust-train & 6.25\% (7) & $+3.57\%\,\uparrow$ \\  
    \midrule
    \multirow{4}{*}{Qwen2-VL-7B} & - & 7.14\% (8) & - \\  
    & GeoQA & 5.36\% (6) & $-1.78\%\,\downarrow$ \\
    & GeoTrust-train & 9.82\% (11) & $+2.68\%\,\uparrow$ \\
    & GeoQA+GeoTrust-train & 8.93\% (10) & $+1.79\%\,\uparrow$ \\  
    \bottomrule
	\end{tabular}
	}
	\label{tab:olympiadbench_think_data}
    \vspace{-0.2em}
\end{table}

\subsubsection{Ablation Study of Thinking Template}

As established in the preceding sections, our \textit{GeoTrust-train} dataset is composed of two primary components: base data and thinking template (TT) data. To validate the effectiveness of the thinking template data, we conduct a series of ablation studies on the GeoQA and Geometry3K datasets. Following the same methodology outlined in~\cref{sec:ablation_of_tt}, we partition the \textit{GeoTrust-train} data into three distinct configurations for our experiments: without thinking templates (w/o TT), replace thinking templates (Re TT), and full data. Each configuration is then combined with the training sets of GeoQA and Geometry3K, respectively, to fine-tune the models. The models are subsequently evaluated on the corresponding test sets.

\begin{figure}[t]
\centering
\includegraphics[width=0.99\linewidth]{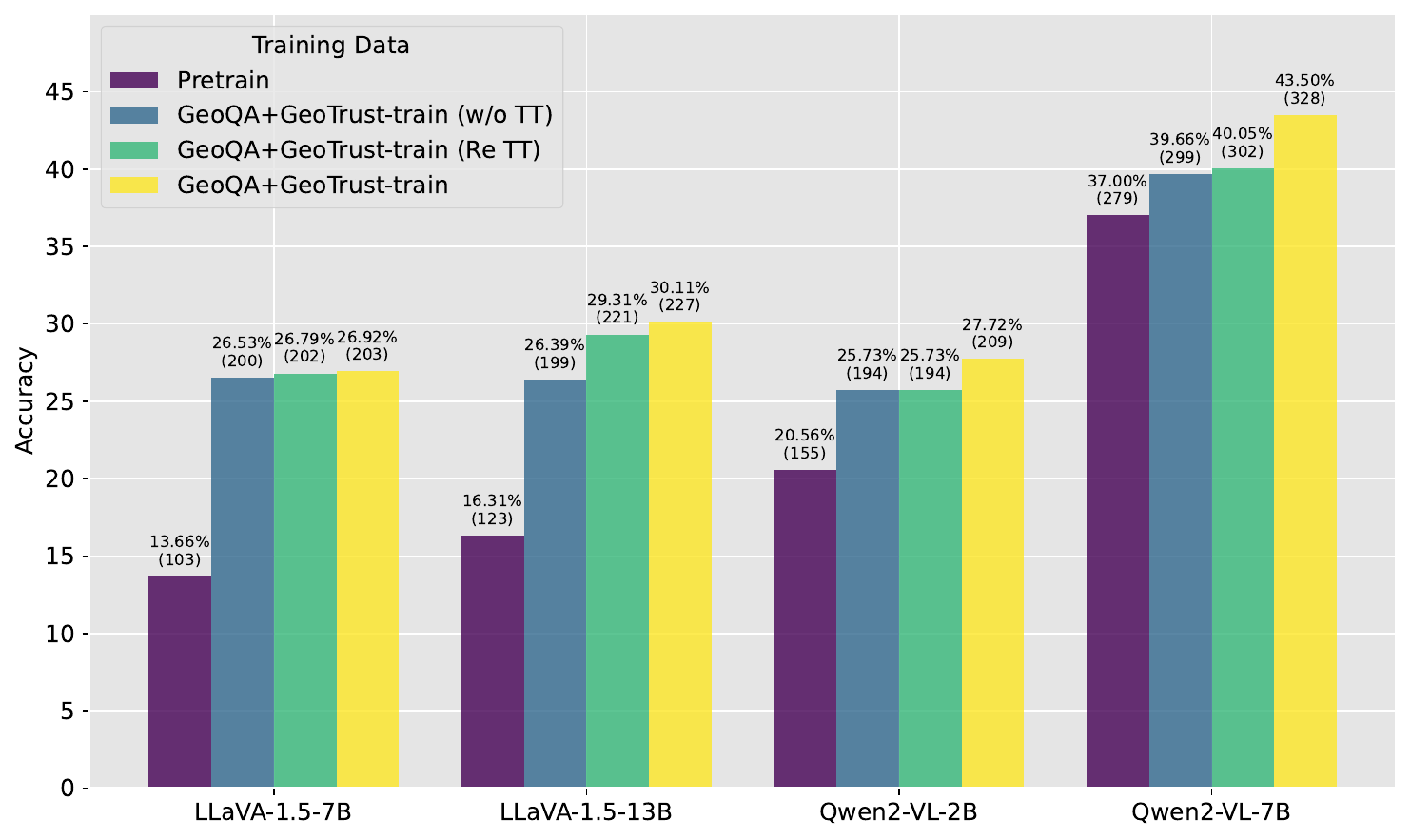}
\caption{Ablation Study of Thinking Template on GeoQA}
\label{fig:tt_ablation_geoqa}
\vspace{-5pt}
\end{figure}

The experimental results on GeoQA, illustrated in~\cref{fig:tt_ablation_geoqa}, reveal that the inclusion of thinking templates consistently enhances performance across all models. For LLaVA-1.5-7B, however, the addition of more data, including thinking templates, yields only marginal improvements. This suggests that the model's performance may have reached a saturation point with the base data alone. In contrast, LLaVA-1.5-13B demonstrates greater sensitivity to data volume; the performance gain from increasing the dataset size is more pronounced than that from introducing thinking templates. Conversely, the Qwen2-VL series benefits more significantly from the addition of thinking templates. This finding aligns with our earlier analysis that since these models are already well-pretrained, they are more responsive to novel data types than to a mere increase in data quantity.

\begin{figure}[t]
\centering
\includegraphics[width=0.99\linewidth]{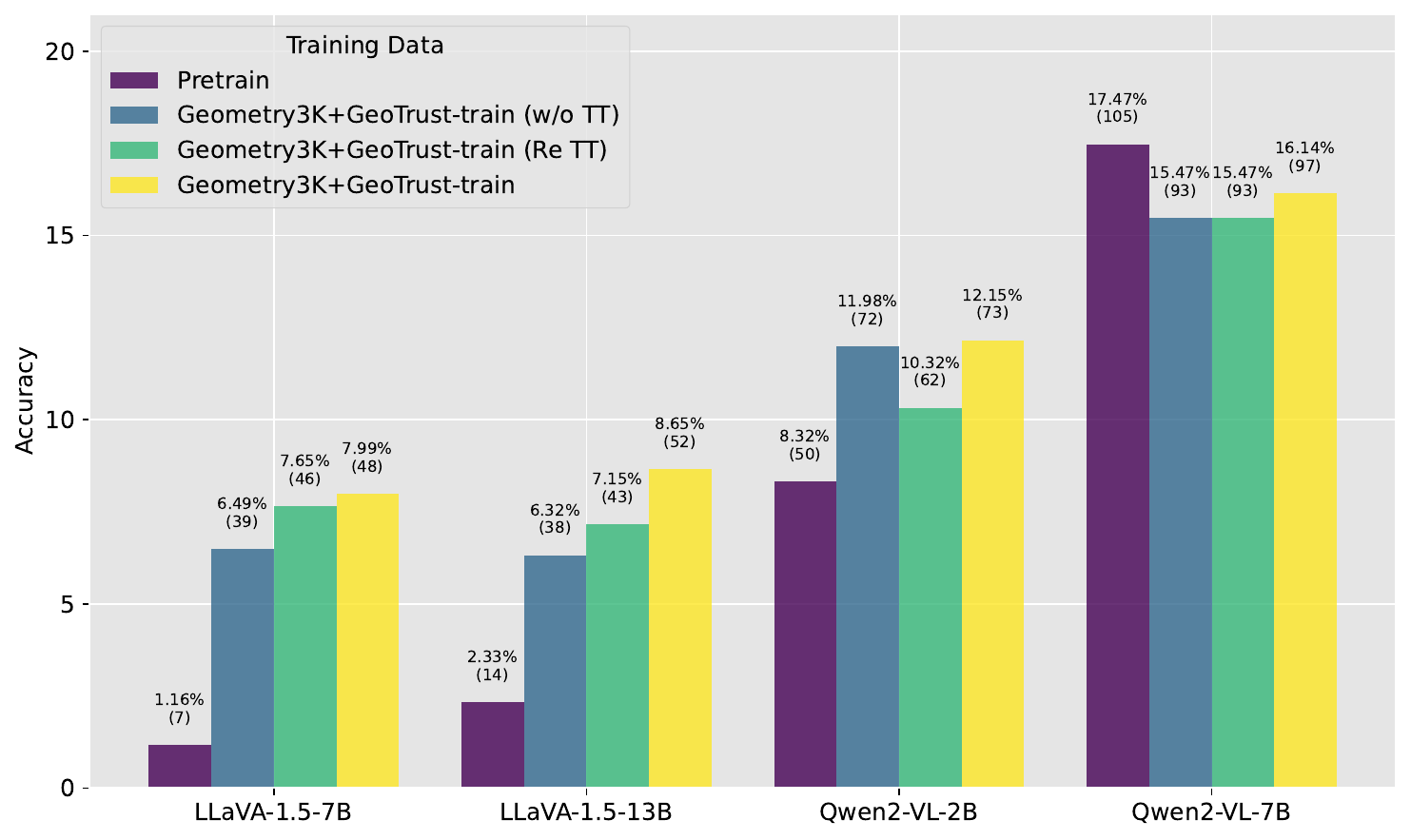}
\caption{Ablation Study of Thinking Template on Geometry3K}
\label{fig:tt_ablation_gemetry3k}
\vspace{-5pt}
\end{figure}

The results on Geometry3K, as depicted in~\cref{fig:tt_ablation_gemetry3k}, present a different pattern. Notably, Qwen2-VL-7B achieves the highest score out-of-the-box (i.e., with its original pre-training), a result we attribute to its pre-training data composition. Nevertheless, when fine-tuned on our mixed datasets, the inclusion of thinking templates still provides a clear performance boost. Overall, the thinking templates demonstrate a more pronounced impact on the performance of the LLaVA-1.5 series models in these experiments.

\subsubsection{Summary of Experiments Results}

Our experiments OOD datasets and benchmarks confirm the generalizability of TrustGeoGen, demonstrating its ability to deliver consistent performance enhancements for models across diverse geometric datasets. A key component, "connection thinking", guarantees the logical coherence and variety of the synthesized data, to the extent that models trained solely on our \textit{GeoTrust-train} dataset achieve notable improvements. The evaluated models display intriguing and divergent behaviors. For instance, the LLaVA-1.5 family—representing an earlier generation of models trained with limited geometric data and less sophisticated methods—shows considerable potential for growth. Our synthetic data consistently and significantly elevates its accuracy. Conversely, the Qwen2-VL series, which leverages state-of-the-art training paradigms and extensive geometric data, is already well-pretrained for these tasks. Nevertheless, enriching its training regimen with \textit{GeoTrust-train} still yields further performance gains.

Furthermore, our novel "thinking templates" introduce structured, multi-faceted problem-solving paradigms previously unseen by these models during pre-training. This explains why even the robustly pre-trained Qwen2-VL models derive clear and objective benefits from the templates' novel reasoning structures. 

In summary, TrustGeoGen-synthesized data has demonstrated robust efficacy across a range of datasets and model architectures. Both of its core mechanisms, "connection thinking" and "thinking templates", have been proven effective. These results offer valuable insights and a compelling direction for future research on leveraging formal engines for automated data curation in training advanced models.

\section{Conclusion and Further Discussion}

\textbf{Conclusion.} We have introduced TrustGeoGen, a multi-modal, integrated, and formal-verified engine for automatically generating fully geometric reasoning data. The resultant \textit{GeoTrust} dataset demonstrates proven effectiveness and generalization capabilities. Furthermore, our benchmark, \textit{GeoTrust-test}, reveals critical limitations in existing multimodal large language models (MLLMs) when handling complex geometric reasoning tasks. Notably, the proposed "connection thinking" bridges the gap between formal and natural reasoning, offering a valuable blueprint for synthesizing model training data with formal engines. In parallel, the "thinking template" mechanism enriches the diversity of the synthetic data, providing greater information gain and further enhancing the models' reasoning abilities. TrustGeoGen represents a crucial first step toward trustworthy geometric problem-solving, establishing a new paradigm for generating rigorous and reliable data.

\noindent \textbf{Further Discussion.} We hope that future research will develop from the following two perspectives:

Trustworthy Geometric Problem Solving: Models should produce verifiable reasoning steps, not merely correct answers. Achieving this demands both reliable data and robust evaluation mechanisms are required. Leveraging the generated data from TrustGeoGen, further study can focus on autoformalization approachs that translate natural language reasoning into formalized steps, enabling automated verification of each deduction’s logical validity. Moreover, TrustGeoGen can dynamically generate evaluation data to prevent prior data leakage during evaluation, ensuring higher reliability in unseen geometric scenarios.

Formal Enhanced Reasoning: TrustGeoGen's formalized reasoning environment enables the generation of trustworthy geometric data by constructing rigorous reasoning graphs that ensure logical correctness at each step. These graphs provide a structured foundation for exploring diverse mathematical reasoning strategies through tailored sampling methods: the current work implements multi-solution and self-reflection traceback data, while future extensions could incorporate the idea of reverse thinking and categorical discussion, etc. These ideas could be further enhanced by integrating alternative training approaches (e.g. RL) to generalize to other reasoning scenarios.

\section*{Acknowledgements}
The research was supported by Shanghai Artificial Intelligence Laboratory, a locally
commissioned task from the Shanghai Municipal Government, National Natural Science Foundation of China (Grant No. 92370201 and 62222607), and the Shanghai Municipal Science and Technology Major Project (Grant No. 22DZ1100102).

\ifCLASSOPTIONcaptionsoff
  \newpage
\fi

\bibliographystyle{IEEEtran}
\bibliography{mybib}



\vfill

\end{document}